\documentclass[journal]{IEEEtran}
\usepackage{cite}

\ifCLASSINFOpdf
  \usepackage[pdftex]{graphicx}
  \graphicspath{{./figures/}}
  \DeclareGraphicsExtensions{.pdf,.jpeg,.png}
\else
  \usepackage[dvips]{graphicx}
  \graphicspath{{./figures/}}
  \DeclareGraphicsExtensions{.eps}
\fi
\usepackage{amsmath}
\def\etal{\emph{et al.~}}
\usepackage{graphicx}
\usepackage{epsfig}
\usepackage{epstopdf}
\usepackage{booktabs}       
\usepackage{amsmath}
\usepackage{multirow}
\usepackage{capt-of}
\usepackage{makecell}

\usepackage[normalem]{ulem}
\usepackage[dvipsnames]{xcolor}

\begin{document}
%
\title{Unsupervised Identification of Disease Marker Candidates in Retinal OCT Imaging Data}
%
%
%

\author{Philipp~Seeb{\"o}ck,
	Sebastian~M.~Waldstein*,
	Sophie~Klimscha,
	Hrvoje~Bogunovic,
	Thomas~Schlegl,
	Bianca~S.~Gerendas,
	Ren{\'e}~Donner,
	Ursula~Schmidt-Erfurth,
	and~Georg~Langs
	\thanks{Copyright (c) 2017 IEEE. Personal use of this material is permitted. However, permission to use this material for any other purposes must be obtained from the IEEE by sending a request to pubs-permissions@ieee.org.}
	\thanks{Manuscript received November,29,2017; revised August,08,2018; accepted October, 13, 2018.}
	\thanks{The financial support by the Christian Doppler Research Association, the Austrian Federal Ministry for Digital and Economic Affairs and the National Foundation for Research, Technology and Development, by the Austrian Science Fund (FWF I2714-B31), and by IBM (2016-2017 IBM PhD Fellowship Award, and Faculty Award) is gratefully acknowledged. A Tesla K40 used for this research was donated by the NVIDIA Corporation.}
	\thanks{P.~Seeb{\"o}ck, T.~Schlegl, R.~Donner and G. Langs are with the Computational Imaging Research Lab, Department of Biomedical Imaging and Image-guided Therapy, Medical University Vienna, Austria  (email: \mbox{philipp.seeboeck@meduniwien.ac.at}, \mbox{georg.langs@meduniwien.ac.at})}
	\thanks{P.~Seeb{\"o}ck, S.M. Waldstein, H.~Bogunovic, S.~Klimscha, B.S. Gerendas, U. Schmidt-Erfurth and G. Langs are with the Christian Doppler Laboratory for Ophthalmic Image Analysis, Vienna Reading Center, Department of Ophthalmology and Optometry, Medical University Vienna, Austria. (email: \mbox{sebastian.waldstein@meduniwien.ac.at})}
	\thanks{* corresponding author}}
	
%
%
%

\markboth{IEEE TRANSACTIONS ON MEDICAL IMAGING,~VOL.~xx,~NO.~xx, OCTOBER 2018}%
{Seeb{\"o}ck \MakeLowercase{\textit{et al.}}: Unsupervised Identification of Disease Marker Candidates in Retinal OCT Imaging Data}%

%



\maketitle

\begin{abstract}
The identification and quantification of markers in medical images is critical for diagnosis, prognosis, and disease management. Supervised machine learning enables the detection and exploitation of findings that are known \textit{a priori} after annotation of training examples by experts. However, supervision does not scale well, due to the amount of necessary training examples, and the limitation of the marker vocabulary to known entities. In this proof-of-concept study, we propose unsupervised identification of anomalies as candidates for markers in retinal Optical Coherence Tomography (OCT) imaging data without a constraint to a priori definitions. We identify and categorize marker candidates occurring frequently in the data, and demonstrate that these markers show predictive value in the task of detecting disease. A careful qualitative analysis of the identified data driven markers reveals how their quantifiable occurrence aligns with our current understanding of disease course, in early- and late age-related macular degeneration (AMD) patients. A multi-scale deep denoising autoencoder is trained on healthy images, and a one-class support vector machine identifies anomalies in new data. Clustering in the anomalies identifies stable categories. Using these markers to classify healthy-, early AMD- and late AMD cases yields an accuracy of 81.40\%. In a second binary classification experiment on a publicly available data set (healthy vs. intermediate AMD) the model achieves an AUC of 0.944.
\end{abstract}


%
\IEEEpeerreviewmaketitle

\section{Introduction}
\label{introduction}
The detection of diagnostically relevant markers in imaging data is critical in medical research and practice. Biomarkers are required to group patients into clinically meaningful subgroups regarding disease, disease progression, or treatment response. Imaging data provides a wealth of information relevant for this grouping in the form of \emph{imaging biomarkers}. Typically, image analysis methods are trained based on \textit{a priori} defined categories, and annotated imaging data. This makes large-scale annotation necessary, which may be costly or infeasible, limits detection to known marker categories, and, overall, slows down the discovery of novel markers.
In contrast, unsupervised detection of anomalies and subsequent data-driven identification of new markers offer the possibility for unbiased classification of a disease and the identification of novel risk factors.
Unsupervised detection can extend our knowledge about the underlying pathophysiology of diseases. The resulting biomarkers can enable a description of the entire spectrum of a disease, from the earliest manifestations to the terminal stages~\cite{mayeux2004biomarkers}.
In this proof-of-concept study, we perform anomaly detection on retinal images to identify biomarker candidates, categorize them, and evaluate their link to disease.

\subsection{Clinical background}
\begin{figure*}[t]
	\centering
	\includegraphics[height=3cm]{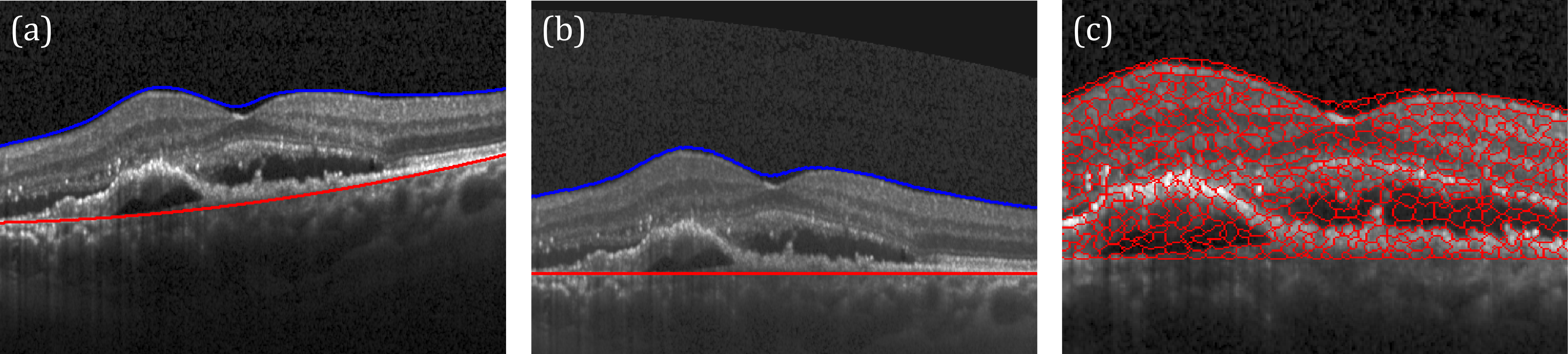}
	\caption{Preprocessing of OCTs. (a) The original OCT scan with the top layer highlighted in blue and the bottom layer (Bruch's membrane) in red. (b) The same scan after normalization was applied (shift to horizontal plane, brightness, contrast). (c) A zoomed-in snippet of the over-segmentation result.}
	\label{fig:method_preprocessing}
\end{figure*}

OCT \cite{huang1991optical} provides high-resolution, 3D volumes of the retina and is the most important diagnostic modality in ophthalmology. Approximately 30 million ophthalmic OCT procedures are conducted per year worldwide, on par with imaging modalities such as magnetic resonance imaging, computed tomography, and positron emission tomography~\cite{fujimoto2016oct}. Each position of the retina sampled by an optical beam results in a vector, the A-scan. Adjacent A-scans form a 2D slice, alias B-scan, which consecutively form the entire volume. Examples of B-scans are shown in Fig.~\ref{fig:method_overallArchitecture} on the left.

Retinal diseases causing vision loss affect many patients. For instance, age-related macular degeneration (AMD) is the leading cause of blindness in industrialized countries and has a worldwide prevalence of 9\% \cite{wong2014global}. Even-though intraretinal fluid shows some predictive value~\cite{waldstein2016correlation}, we are lacking accurate and reliable imaging markers and predictors for individual patients disease courses. The discovery of novel reliable markers in imaging data is relevant to enhance individual care, encompassing the identification and categorization of marker candidates, and the quantification of their link to disease. Not all patterns occurring in OCT volumes are understood or interpretable, and for certain retinal diseases such as for AMD~\cite{SchmidtErfurth20161}, pathogenic mechanisms are not yet fully known.

Computational anomaly detection \cite{pimentel2014review} and categorization is a natural approach to tackle this problem, where the former is defined as the detection of cases that differ from the normal samples available during training. In retinal images this is a difficult task for many reasons.
In contrast to natural images, such as photographs, in retinal imaging the findings relevant for diagnosis cover only a small fraction of the overall volume. Furthermore, their deviation from normal tissue is subtle compared to the variability of healthy retinas.
Therefore, to identify novel marker candidates, we form a model of normal tissue variability, and detect anomalies deviating from this model.
In this paper, we define \emph{normal} as the absence of pathological changes beyond age-related alterations, i.e. the only allowed visible alteration included drusen below 63 $\mu m$ in size according to the Beckman Initiative Classification~\cite{ferris2013clinical}. This definition accounts for age-related changes that normally do not result in visual impairment. For instance, the majority of elderly patients shows small drusen, while still maintaining normal vision.

Some retinal diseases such as retinal vein occlusion (RVO), often occur unilaterally. Thus, the contralateral eye is not affected by the acute event of the disease and can be elegantly used as training data for the normal appearance model. In our case, these volumes of contralateral eyes were screened by a retinal specialist to rule out cases with pathological changes.
Since our model is purely trained on normal data, we omit the need to collect a dataset containing a sufficient amount of anomalies representing the entirety of their possible variability. At the same time, the applicability of the model is not limited to a specific disease.

\subsection{Related Work}
\label{introduction:relatedwork}
Anomaly detection can be a crucial first step in the process of biomarker detection. The results of these algorithms  are affected by the quality of the features used for characterizing the data.
Supervised deep learning has recently improved the state-of-the-art in various tasks, such as image classification \cite{szegedy2015going}, object detection \cite{ren2015faster} or weakly supervised learning linking semantic descriptions to image content~\cite{schlegl2015automatic}. It results in rich feature representations, although at the cost of requiring large amounts of annotated training samples and the limitation to known markers.
On the other hand, unsupervised learning enables the exploitation of unlabeled data, capturing the structure of its underlying distribution~\cite{doersch2015unsupervised,dosovitskiy2014discriminative,zhao2015stacked,zhou2015joint}. A well-known and widely used technique for feature learning is Principal Component Analysis (PCA) \cite{lopez2011principal}, which is computationally efficient, but limited to a linear embedding. In contrast, unsupervised deep learning of convolutional neural networks (CNNs) is computationally more expensive, but can learn a non-linear embedding.

In \cite{dosovitskiy2014discriminative}, unsupervised CNN training was performed by discriminating between surrogate image classes created by manually defined data augmentation to render the resulting representation robust to certain transformations. Other studies propose incorporation of supervisory signals such as spatial context \cite{doersch2015unsupervised} to omit the requirement of manually annotated data. In our study we identify clinically relevant biomarkers without prior human input, which could bias the result.

The proposed anomaly detection method is inspired by the combination of Deep Belief Networks (DBNs) with One-Class SVMs for anomaly detection in real-life datasets, which have considerably different characteristics compared to medical images~\cite{erfani2016high}. The DBNs learn a feature representation, while the One-Class SVM finds a boundary describing regions in the feature space with high probability density of the training data.
Erfani \etal~\cite{erfani2016high} trained DBNs in a layer-wise fashion, and did not use a multi-scale architecture, as we did.
According to \cite{zhao2015stacked,zhou2015joint}, the combination of joint training of layers and local regularization constraints for each layer is more advantageous than layer-wise training without constraints. Therefore, we both trained the deep convolutional autoencoder (DCAE)~\cite{zhao2015stacked} and the deep denoising autoencoder (DDAE)~\cite{zhou2015joint} jointly. The multi-scale architecture was partly inspired by~\cite{schlegl2015automatic}, where weakly supervised learning was used to link image information to semantic descriptions of image content.
Since DCAEs are specifically designed to learn effective representation of images, they are a logical comparison method when learning unsupervised image representations.
The idea of using normal subjects to model a normal population is not novel.
Sibide \etal \cite{sidibe2017anomaly} modeled the appearance of normal OCT B-scans with a Gaussian Mixture Model (GMM) and detected anomalous B-scans as outliers. The number of outliers served as basis for classification of an entire OCT volume. In contrast, we aim at pixel level anomaly detection.
In \cite{dufour2012pathology}, a shape model of normal retinal layers is used to segment anomalies. The model has a close fit in normal regions, while there is no fit in areas of anomalous shapes. The limitations of this approach are that it heavily depends on the quality of the layer segmentation algorithm and does not take into account image information explicitly to detect anomalies.
Finally, Schlegl \etal \cite{schlegl2017unsupervised} proposed AnoGAN, a deep convolutional Generative Adversarial Network (GAN) to learn a manifold of normal anatomical variability, in order to identify anomalous regions in OCT images. AnoGAN is restricted to healthy representations by definition, which makes it inappropriate for a straightforward subsequent clustering step of anomalies. In contrast, we aim at learning a feature representation with our autoencoder approach which is general enough to enable a meaningful embedding of anomalies, though we solely need normal training data.

Examples for the classic biomarker identification strategy are \cite{farsiu2014quantitative,lad2018evaluation}, where the authors used a priori defined features in a supervised way to evaluate the applicability as biomarkers for specific diseases. In contrary, our approach focuses on identifying new marker candidates in an unsupervised way.

Regarding classification of retinal diseases on volume level, main focus of related work \cite{farsiu2014quantitative,venhuizen2015automated} is to solve the classification task itself. In contrast, here our target is to evaluate the link of identified categories, alias marker candidates, to disease by using them as features for classification.

\subsection{Contribution}
We propose a method to identify marker candidates in imaging data in an unsupervised fashion. Our approach first separates anomalous candidates from normal tissue in retinal spectral-domain Optical Coherence Tomography (SD-OCT) based on the features learned by \emph{DDAE} on healthy samples, and a One-Class SVM to model normal appearance distribution. We identify categories of frequently occurring anomalies using clustering, and evaluate their link to disease. In a qualitative evaluation, retinal experts could map part of the categories identified by this approach to known retinal structures.
At the same time other categories remain as novel anomaly candidates, for which results on the classification tasks suggest that they are also linked to disease.

This paper is an extension of our previous work~\cite{seebock2016identifying} introducing a new feature-learning approach, and more in-depth evaluation of anomaly detection, categorization, and the link of these marker candidates to disease.

\section{Methods}
\label{method}
\begin{figure*}[t]
	\centering
	\includegraphics[clip,trim=1.5cm 5.6cm 2cm 4.8cm,height=6.4cm]{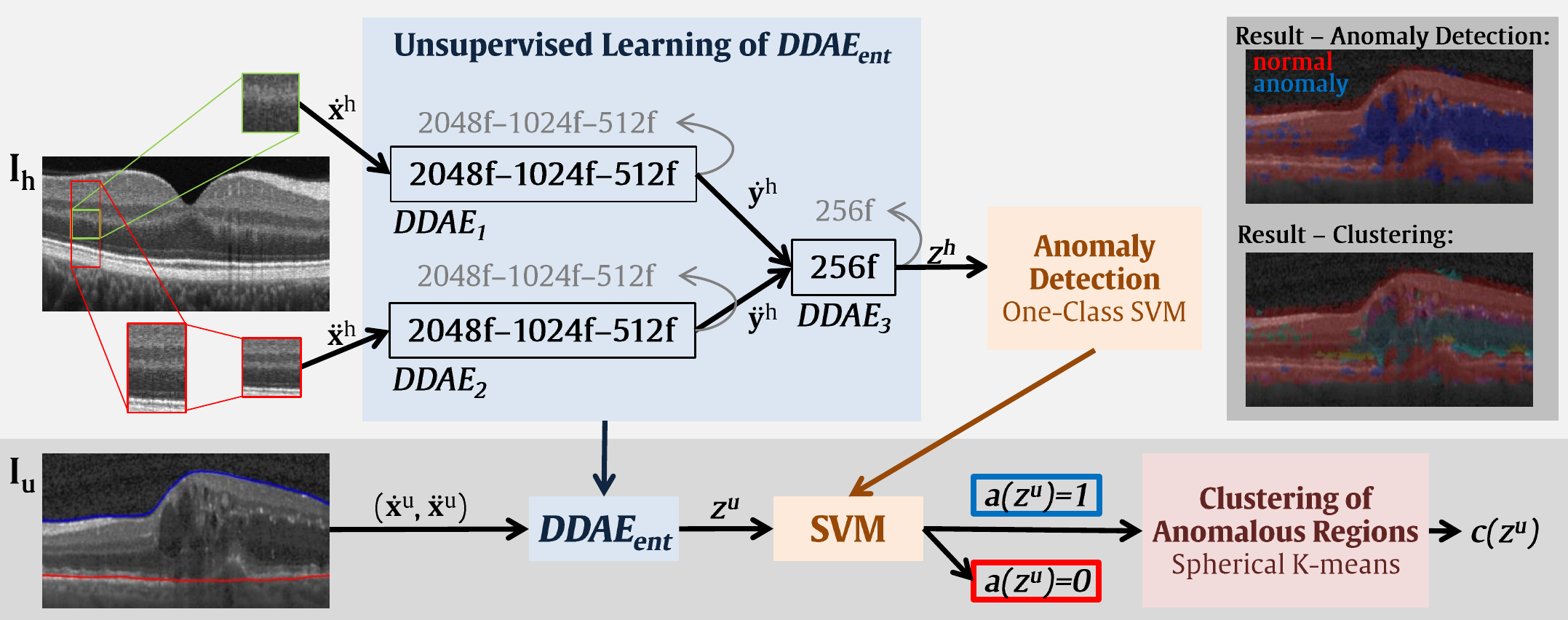}
	\caption{Multi-scale architecture used in the experiments. Pairs of $\dot{x}_i^h = 32\times32$ (green) and $128\times32$ (red) patches were extracted at positions $p_i^h$ in healthy OCTs $I_h$ from B-scans, where the larger patches were downsampled to $\ddot{x}_i^h = 32\times32$, as illustrated on the left. The encoders of all unsupervised learning modules are denoted in black, while the decoders are depicted in gray. Once learned, the encoders of $ DDAE_{ent} $ were used to create the new feature representation $ z_i^h $, for a specific superpixel $s_i^h$, with position $p_i^h$ in the B-scan. Subsequent training of a One-Class SVM enabled the opportunity to detect anomalous regions, $a(s_i^u)=1$, in unseen B-scans $I_u$~(i.e. B-Scans not used during training). In order to subdivide anomalous regions into meaningful categories, $c_j$, clustering was performed. This means that for each superpixel $s_i^u$ with position $p_i^u$ in an anomalous region, a cluster assignment $c(s_i^u)$ was performed. An example of anomaly detection and subsequent clustering of anomalous regions is shown on the top right.}
	\label{fig:method_overallArchitecture}
\end{figure*}

To capture visual information at different levels of detail, we used a multi-scale approach to perform superpixel-wise segmentation of the visual input. While the preprocessing steps are shown in Fig.~\ref{fig:method_preprocessing}, the overall architecture is illustrated in Fig.~\ref{fig:method_overallArchitecture}.
After preprocessing (Section~\ref{method:preprocess}), 2D-patches extracted from B-Scans from healthy OCT volumes were used to train a deep denoising autoencoder model (Section \ref{method:unsupervised}), which provided an embedding that represented healthy anatomical variability. A One-Class SVM was trained on this embedding to obtain a boundary, which encompassed the distribution of healthy patches (Section~\ref{method:anomaly}). Using this boundary, unseen volumes~(i.e. volumes not used during training) were segmented into healthy or anomalous regions. Subsequent clustering of anomalous regions partitioned anomalies into more specific categories (Section~\ref{method:clustering}).

\subsection{OCT Preprocessing}
\label{method:preprocess}
For all OCT volumes, we identified the top (Internal Limiting Membrane - ILM) and bottom (Bruch's Membrane - BM) layer of the retina using a graph-based surface segmentation algorithm \cite{garvin2009automated}, where the bottom layer was used to flatten the retina by projecting it to a horizontal plane, as shown in Fig.~\ref{fig:method_preprocessing}(b). The top and bottom layer of the retina are also illustrated as blue and red in Fig.~\ref{fig:method_overallArchitecture} at the bottom left. This reduced the differences in appearance caused by varying orientations and positions of the retina within the volume. We applied brightness and contrast normalization for each B-scan and added a constant to shift the values into a positive range. The latter was necessary to ensure that the deep denoising autoencoders ($ DDAE_{1} $, $ DDAE_{2} $) were able to reconstruct the input patches ($\dot{x}$, $\ddot{x}$) properly during training. Finally, we performed over-segmentation of B-scans to monoSLIC superpixels, $s$, of an average size of $4\times4$ pixels \cite{mholzer2014superpixel}, as illustrated in Fig.~\ref{fig:method_preprocessing}(c). This merges pixels into homogeneous groups of superpixels, which allows to perform the computations on a reduced number of superpixels as opposed to computations on every pixel.

Preprocessing of healthy B-scans, $I_h$, with $h=1,\dots , H$, resulted in $S^h$ superpixels, $s_i^h$, for each (as illustrated in Fig.~\ref{fig:method_preprocessing}), with center positions $p_i^h$ and $i=1,\dots,S^h$, where $H$ denotes the number of healthy B-Scans, $S^h$ the number of superpixels per B-Scan, $i$ the index of the superpixel, and $h$ the index of the healthy B-Scan.

\subsection{Unsupervised Learning of Appearance Descriptors}
\label{method:unsupervised}
The network architecture of the deep denoising autoencoder consists of an encoding and decoding part. We chose three fully connected layers, with 2048 neurons in the first, 1024 in the second, and 512 in the third layer to build the encoder, with the structure also denoted as \texttt{2048f-1024f-512f}. The mirrored version of the encoder (\texttt{512f-1024f-2048f}) formed the decoder, as illustrated in Fig.~\ref{fig:method_overallArchitecture}. The weight matrices of two corresponding layers were tied: $W_{enc} = W_{dec}^{T}  $. All layers were followed by Exponential Linear Units (ELUs) \cite{clevert2015fast}, with $ \alpha = 1 $:
\begin{equation}
f(x) = 
\begin{cases}
x & \text{if}\ x > 0 \\
\alpha(exp(x) - 1) & \text{if}\ x \le 0
\end{cases}
\end{equation}

The Mean Squared Error function, $ MSE(x, \hat{x}) $, was chosen as a loss function for training, where $ x $ denotes the input patch and $ \hat{x} $ the output of the last layer of the decoder. The autoencoder was trained jointly in an end-to-end fashion, as proposed in \cite{zhou2015joint}. In addition, we added a local constraint to each layer by corrupting the input of every layer in the encoder. More precisely, a fraction of the inputs was set to zero. As opposed to layer-wise training, this corresponds to unsupervised joint training with local constraints in each layer.

We conducted unsupervised training of two deep denoising autoencoders ($ DDAE_{1} $, $ DDAE_{2} $) on the patches, $\dot{x}_i^h$ and $\ddot{x}_i^h$, extracted at center positions of superpixels $p_i^h$ from the healthy B-scans, $I_h$. While $\dot{x}_i^h = 32\times32$ served as input for $ DDAE_{1} $, $ DDAE_{2} $ was trained with $128\times32$ patches $\ddot{x}_i^h$, downsampled to $32\times32$. The provided patch sizes are given in pixels.
Both models were fixed for the subsequent training of another denoising autencoder, $ DDAE_{3} $, its single-layer architecture denoted as \texttt{256f}, with the concatenated feature vectors $[\dot{y} \ddot{y}]$ as input, where $\dot{y}_i^h = DDAE_{1}(\dot{x}_i^h)$ and $\ddot{y}_i^h = DDAE_{2}(\ddot{x}_i^h)$. All three learned encoders from $DDAE_{1}$, $DDAE_{2}$, and $DDAE_{3}$ together formed the final model, $ DDAE_{ent} $, that gave us a 256 dimensional feature representation, $ z_i^h = DDAE_{ent}(\dot{x}_i^h, \ddot{x}_i^h) = DDAE_{ent}(s_i^h)$, for a specific superpixel, $s_i^h$, with corresponding patches $(\dot{x}_i^h, \ddot{x}_i^h)$ extracted at the central position of the superpixel, $p_i^h$. The multi-scale architecture allows to incorporate the local information of the smaller patch and at the same time the neighborhood and orientation information of the larger patch.

\subsection{Anomaly Detection with One-Class SVM}
\label{method:anomaly}
Based on the learned feature representation, $z_i^h$, we estimated the distribution of healthy examples with a One-Class SVM \cite{scholkopf2001estimating}, using a linear kernel. The SVM searches for a boundary that describes the distribution of normal data, which serves as a decision boundary for unseen data. New samples can then be classified either as coming from the same data distribution if lying inside the boundary ($0$, normal) or not ($1$, anomaly).
Since we used a linear kernel for One-Class SVM, the only hyper-parameter was $\nu$. This parameter determines the amount of normal training data that must lie within the boundary, i.e., which is detected as normal. For example, a value of $0.1$ means that 90\% of the training samples are within the boundary. In this work, we chose the parameter value with the highest dice score on the validation set for the final model.

For unseen B-Scans, $I_u$, with $u=1,\dots,U$, features $z_i^u$ and the corresponding class $a(z_i^u) = \{0,1\}$ were computed for each superpixel, $s_i^u$, with position $p_i^u$ within the top and bottom layer of the retina, where $U$ denotes the number of unseen B-Scans. The computed class label $a(z_i^u)$ was assigned to the entire superpixel: $a(z_i^u) = a(s_i^u)$. This provided a segmentation of the retina into two classes at superpixel level.

\subsection{Categorization of Anomalous Regions}
\label{method:clustering}
We used spherical K-means clustering~\cite{hornik2012spherical} with cosine distance to sub-segment anomalous superpixels $a(s_i^u)=1$, which have been classified as anomalous by our method in the first stage, in unseen B-Scans into $C$ clusters, $c(s_i^u) = c(z_i^u) = j $, with $ j = 1,\dots,C $.
More precisely, we trained a cluster model using the 256 dimensional feature representation $z$ on an ''anomaly training set'', that was composed of samples with $a(z_i^u)=1$ only, to obtain cluster centroids $c_j$.
The number of cluster centroids, $C$, was determined by an internal evaluation criterion called the Davies-Bouldin (DB) index~\cite{halkidi2001clustering}, calculated on the anomaly training set. A small value indicates compact and well-separated clusters, hence, the model with the smallest DB index was selected. 

To segment an unseen B-Scan $I_u$, each superpixel with the property $a(s_i^u)=1$ got a cluster assignment, $c(s_i^u)$, where $c(s_i^u)$ gives the index, $j$, of the nearest cluster centroid, $c_j$.  To facilitate reading, we omitted indices $i$, $h$, and $u$ henceforth.

\section{Evaluation}
\label{evaluation}
Our evaluation tests: (1) if the proposed algorithm can identify anomalous regions in imaging data, (2) if the algorithm can detect stable categories of anomalies, and (3) if these categories can serve as disease markers.

\begin{figure*}[t]
	\centering
	\begin{center}
		\setlength{\tabcolsep}{2pt}
		\begin{tabular}{r c c c c}
			\scriptsize{Original} & 
			\raisebox{-0.5\totalheight}{\includegraphics[height=2cm,width=3.9cm]{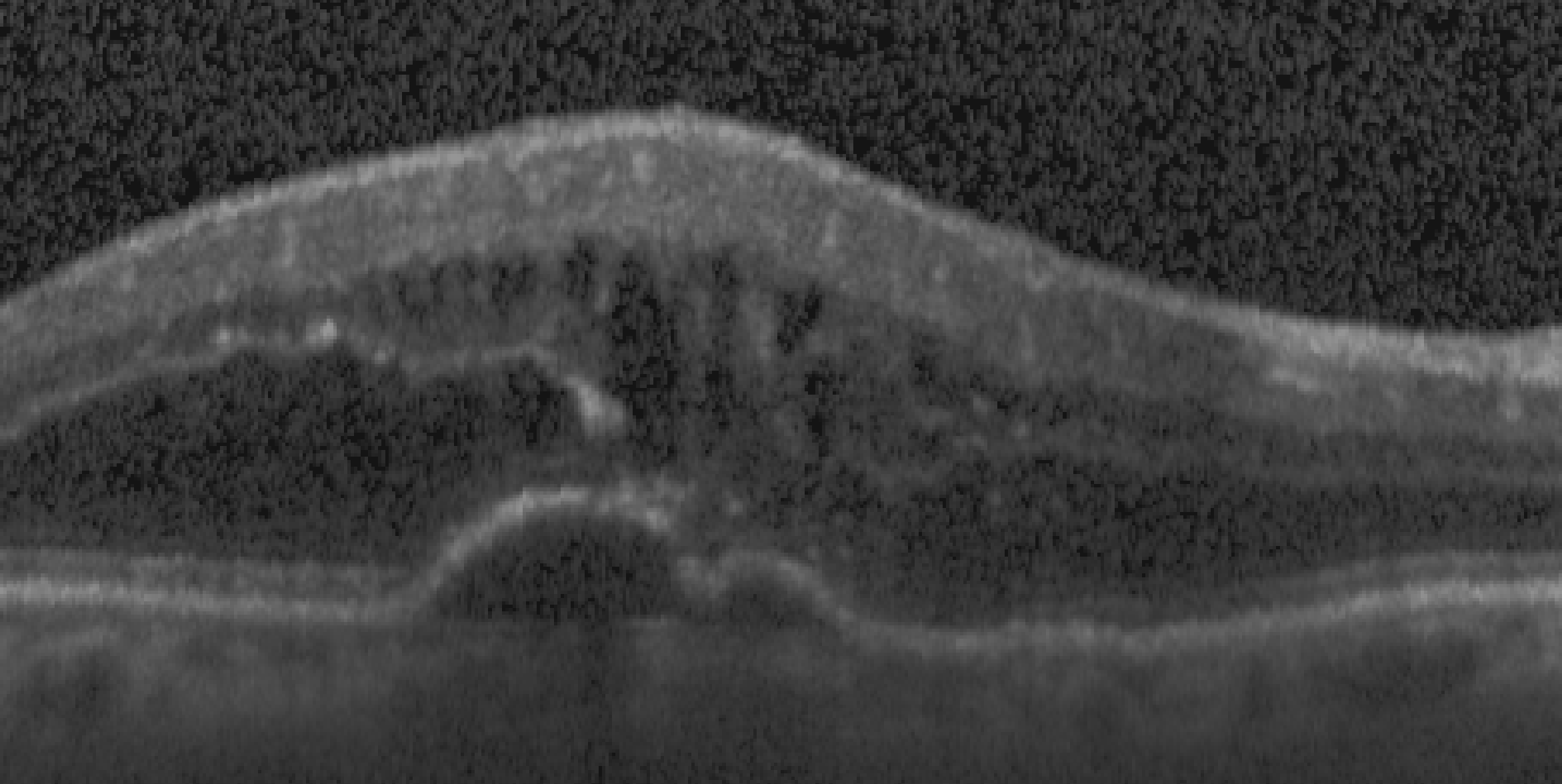}} &
			\raisebox{-0.5\totalheight}{\includegraphics[height=2cm,width=3.9cm]{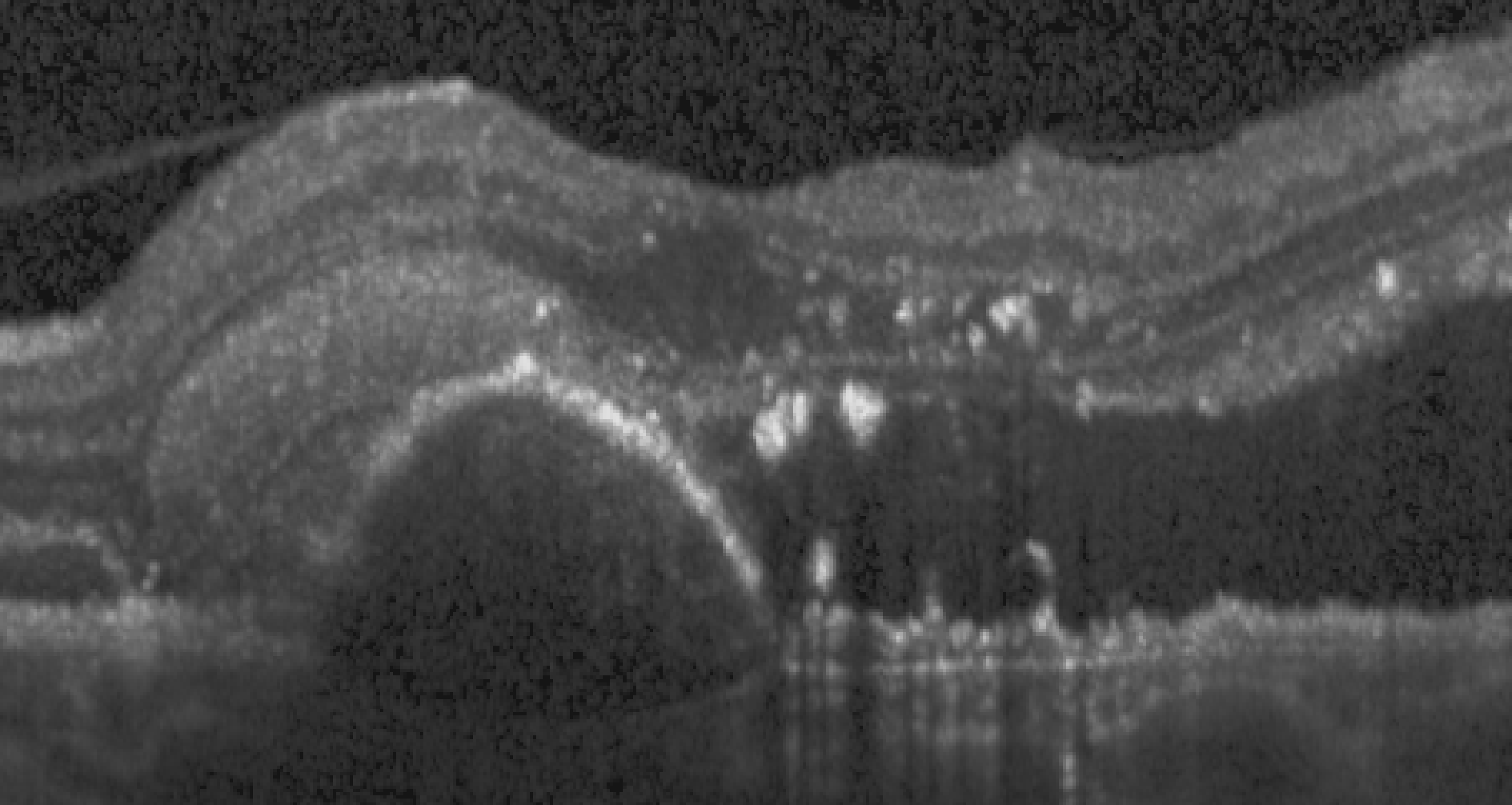}} &
			\raisebox{-0.5\totalheight}{\includegraphics[height=2cm,width=3.9cm]{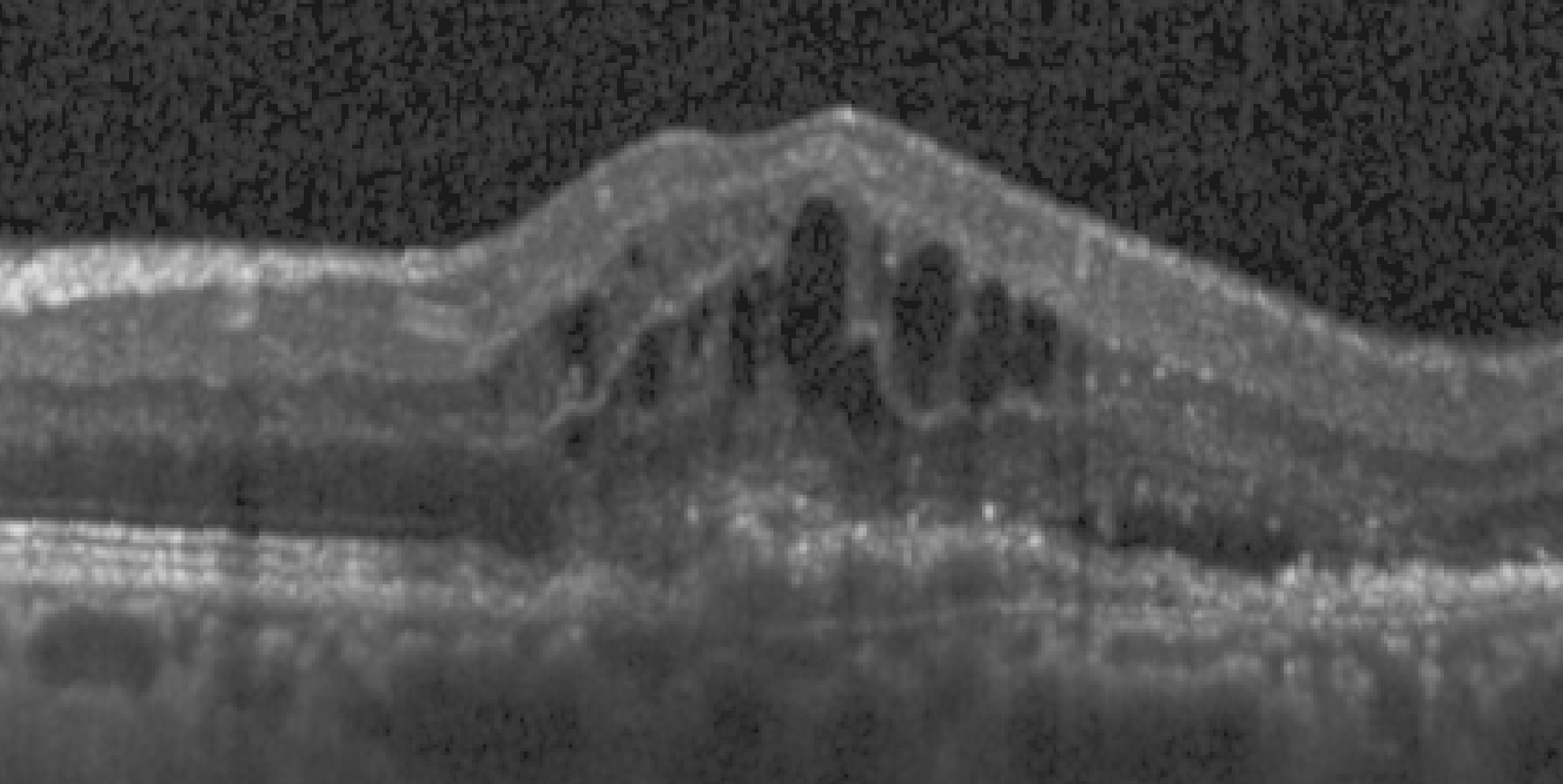}} &
			\raisebox{-0.5\totalheight}{\includegraphics[height=2cm,width=3.9cm]{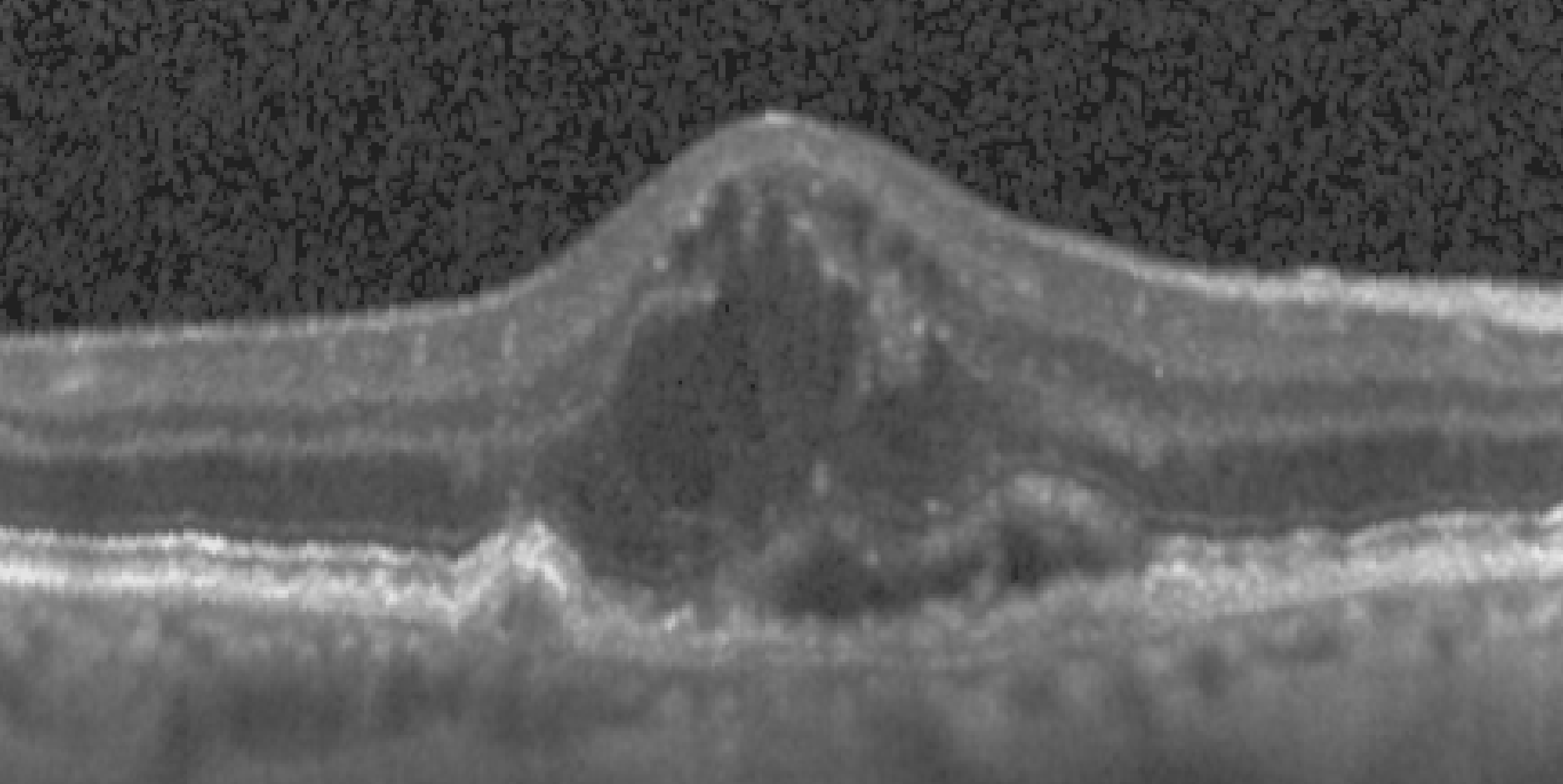}}
			\vspace{3pt} \\
			\scriptsize{Ground Truth} &
			\raisebox{-0.5\totalheight}{\includegraphics[height=2cm,width=3.9cm]{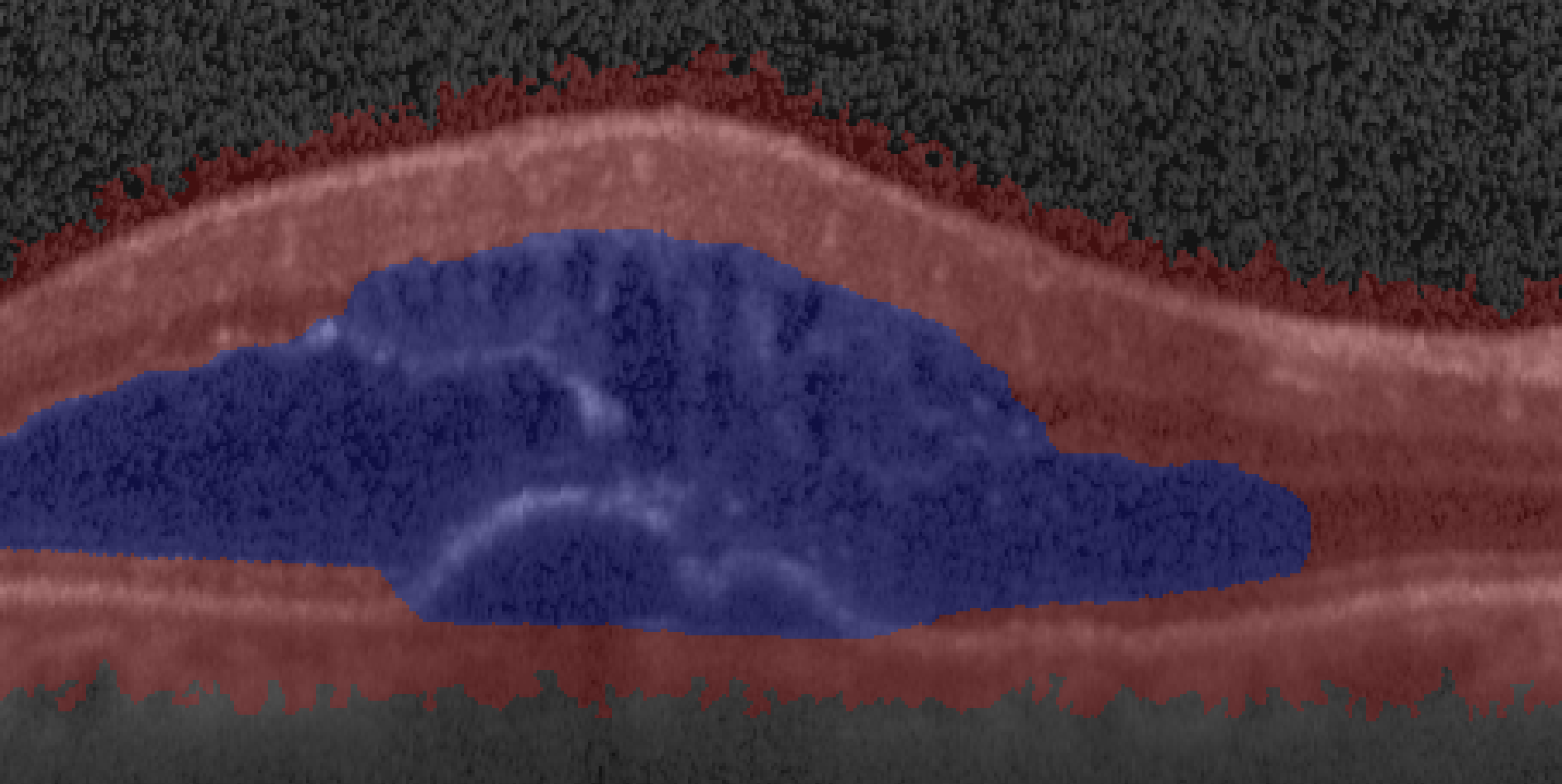}} &
			\raisebox{-0.5\totalheight}{\includegraphics[height=2cm,width=3.9cm]{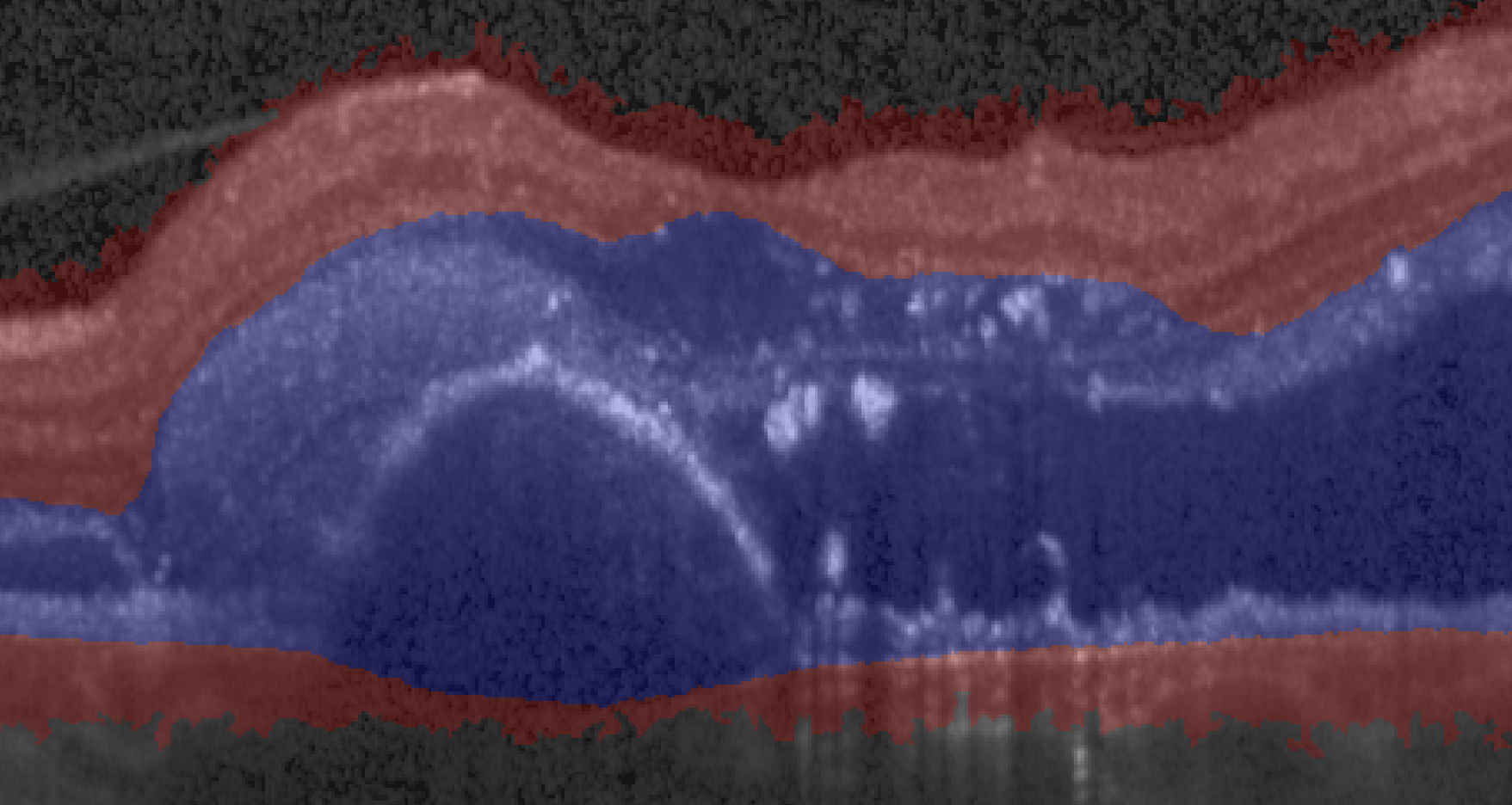}} &
			\raisebox{-0.5\totalheight}{\includegraphics[height=2cm,width=3.9cm]{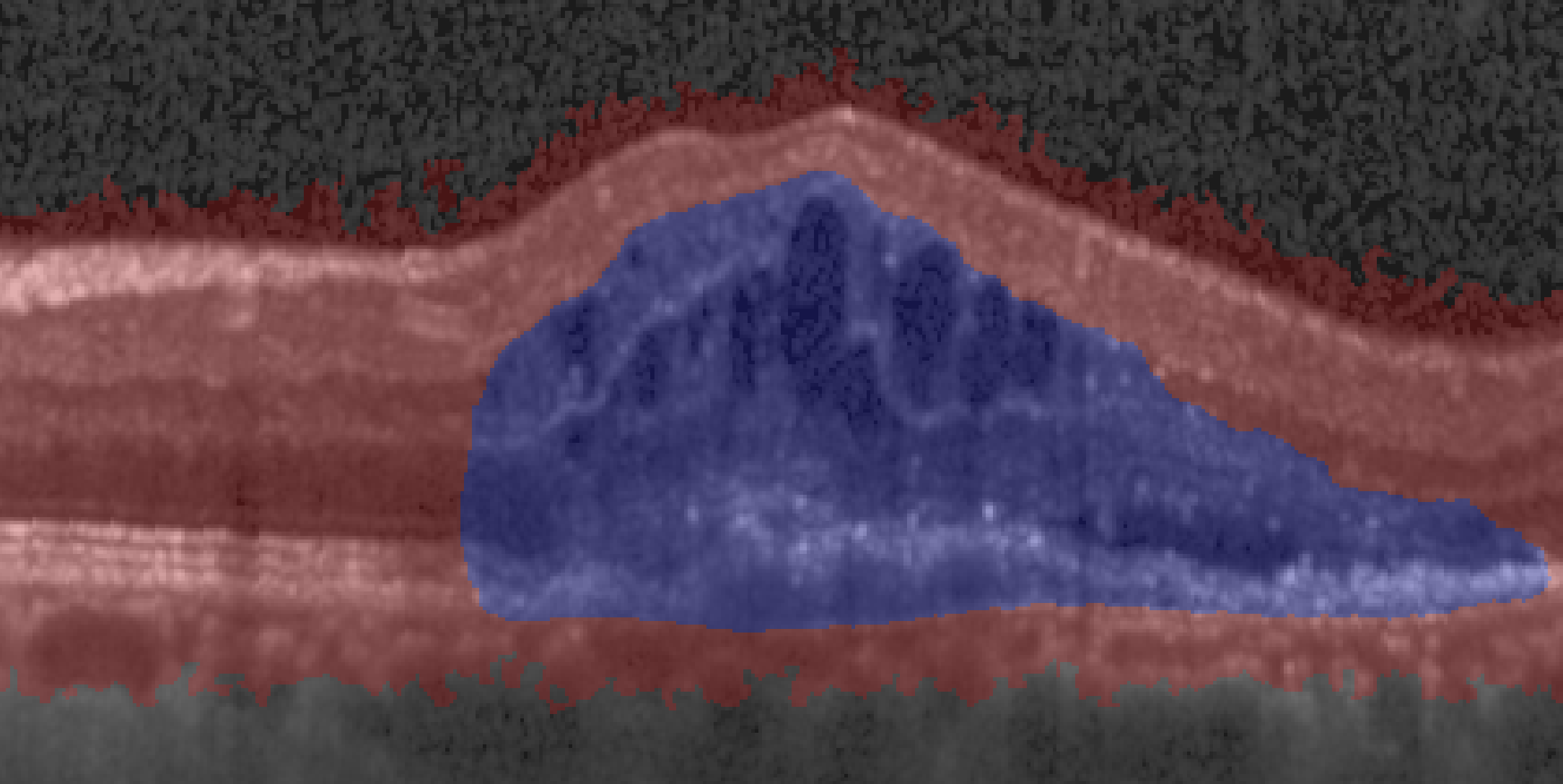}} &
			\raisebox{-0.5\totalheight}{\includegraphics[height=2cm,width=3.9cm]{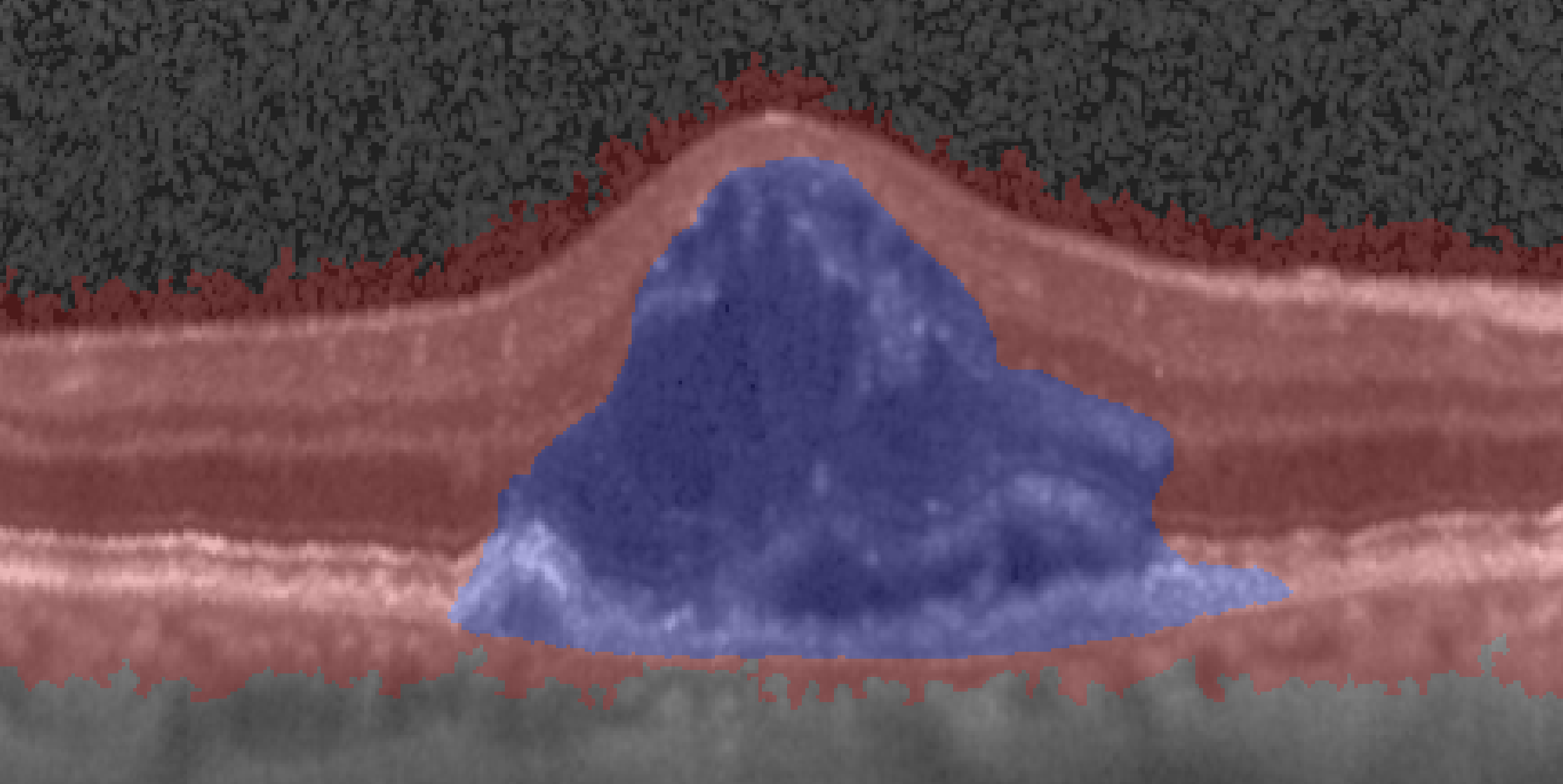}}
			\vspace{3pt} \\
			\makecell[r]{ \scriptsize{$PCA_{256}$} \\ \tiny{\{0.66, 0.63, 0.35, 0.34\}} } & 
			\raisebox{-0.5\totalheight}{\includegraphics[height=2cm,width=3.9cm]{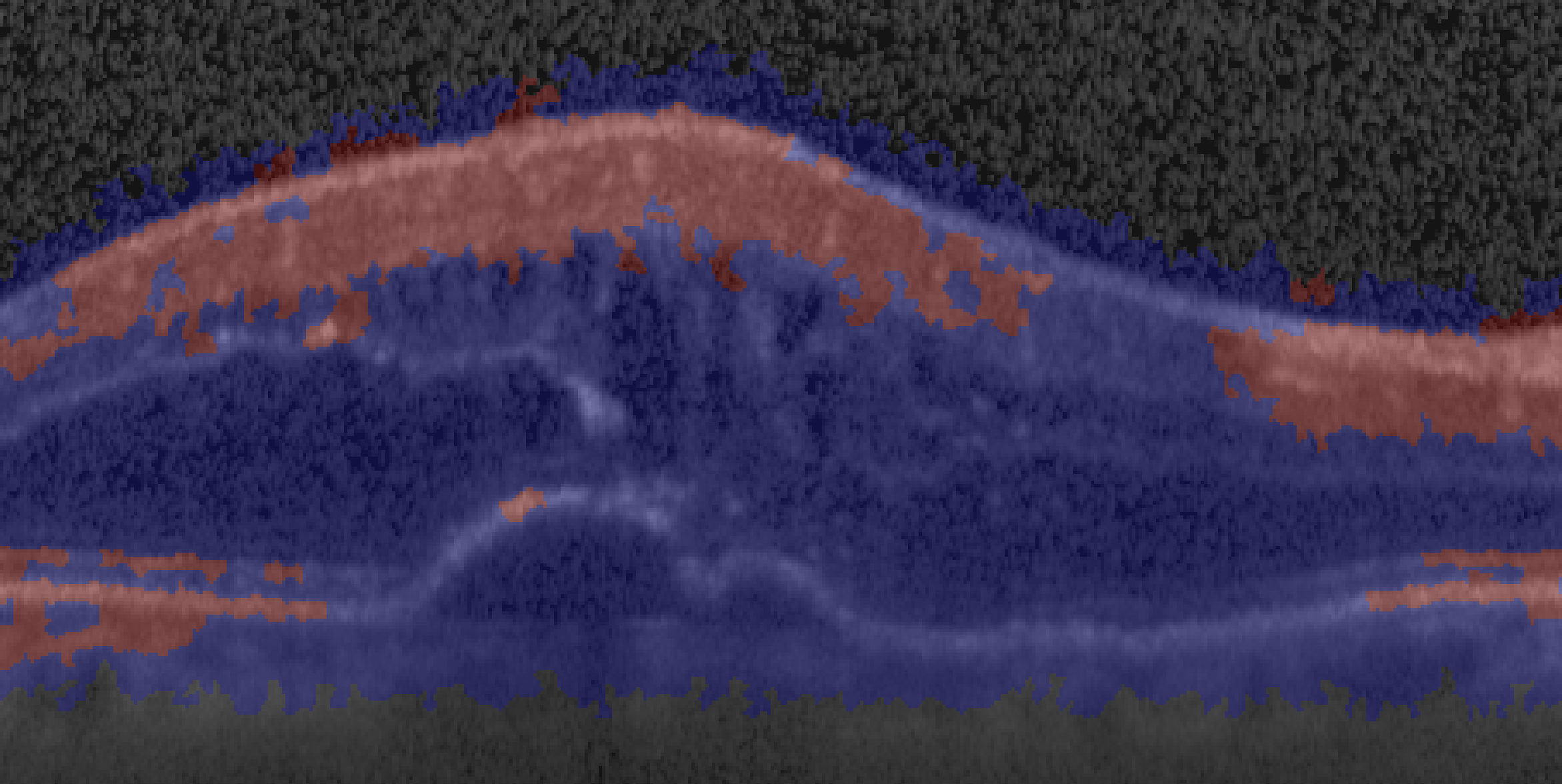}} &
			\raisebox{-0.5\totalheight}{\includegraphics[height=2cm,width=3.9cm]{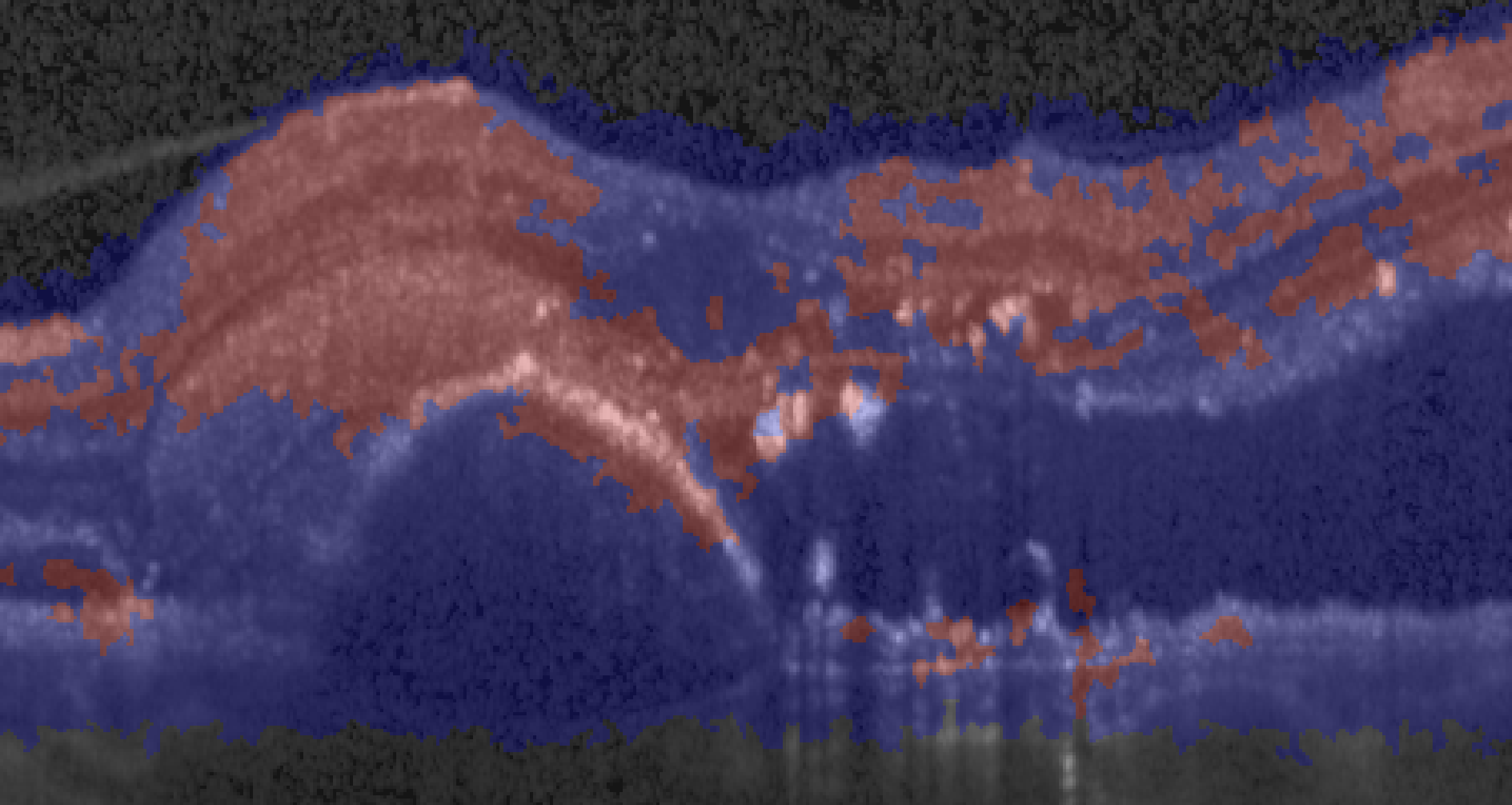}} &
			\raisebox{-0.5\totalheight}{\includegraphics[height=2cm,width=3.9cm]{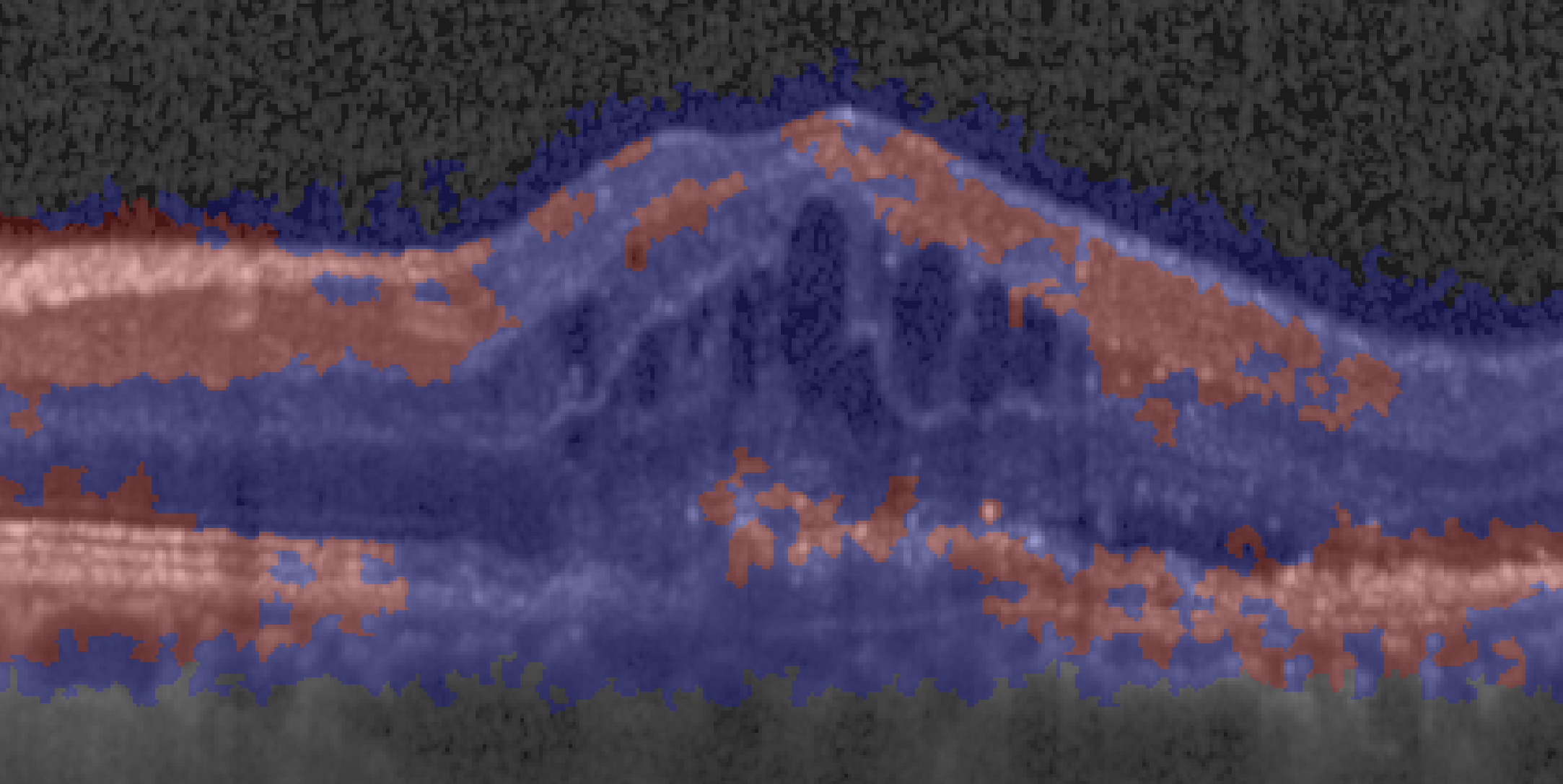}} &
			\raisebox{-0.5\totalheight}{\includegraphics[height=2cm,width=3.9cm]{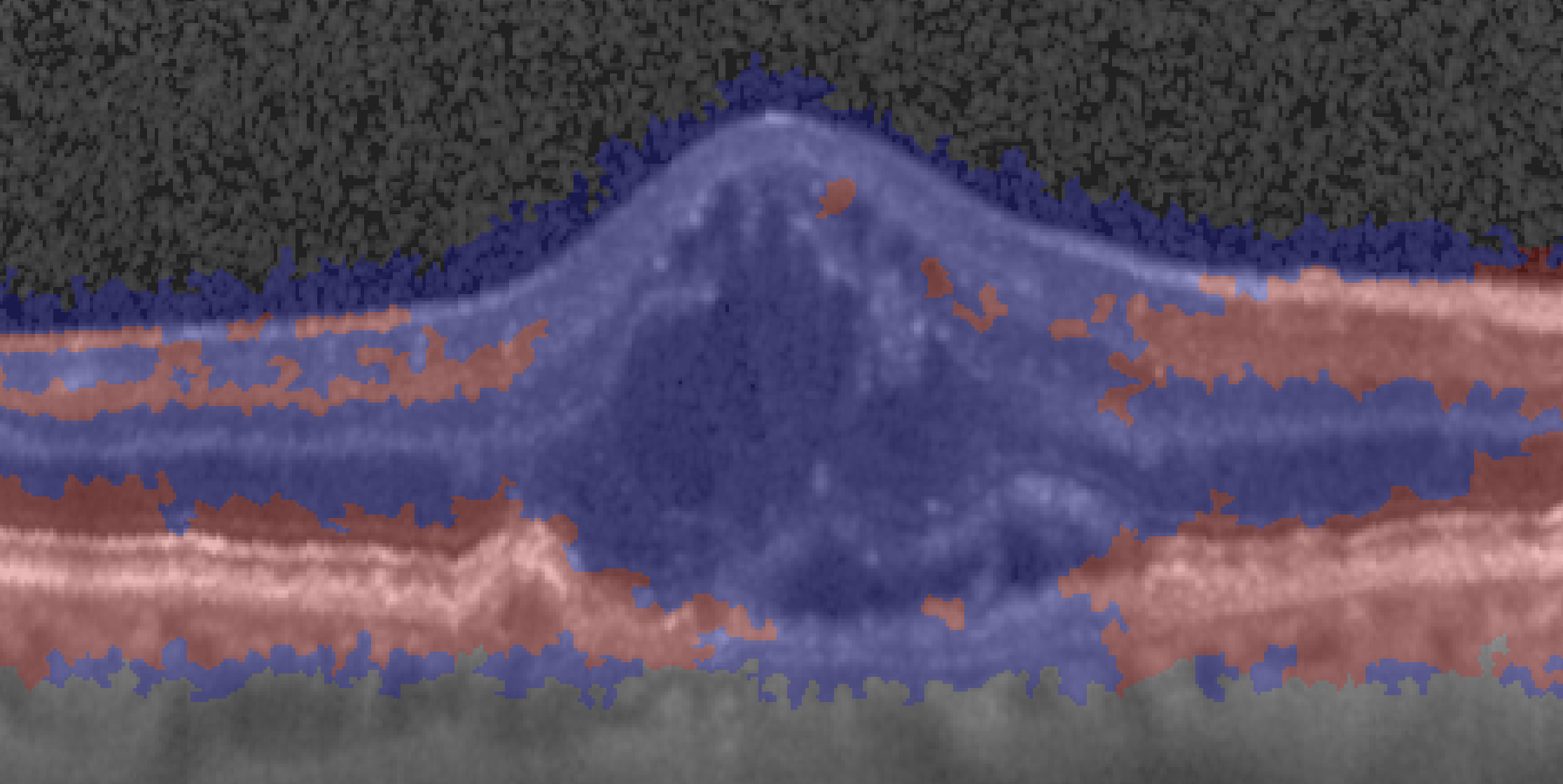}}
			\vspace{3pt} \\
			\makecell[r]{ \scriptsize{$PCA_{0.95}$} \\ \tiny{\{0.64, 0.71, 0.38, 0.32\}} } & 
			\raisebox{-0.5\totalheight}{\includegraphics[height=2cm,width=3.9cm]{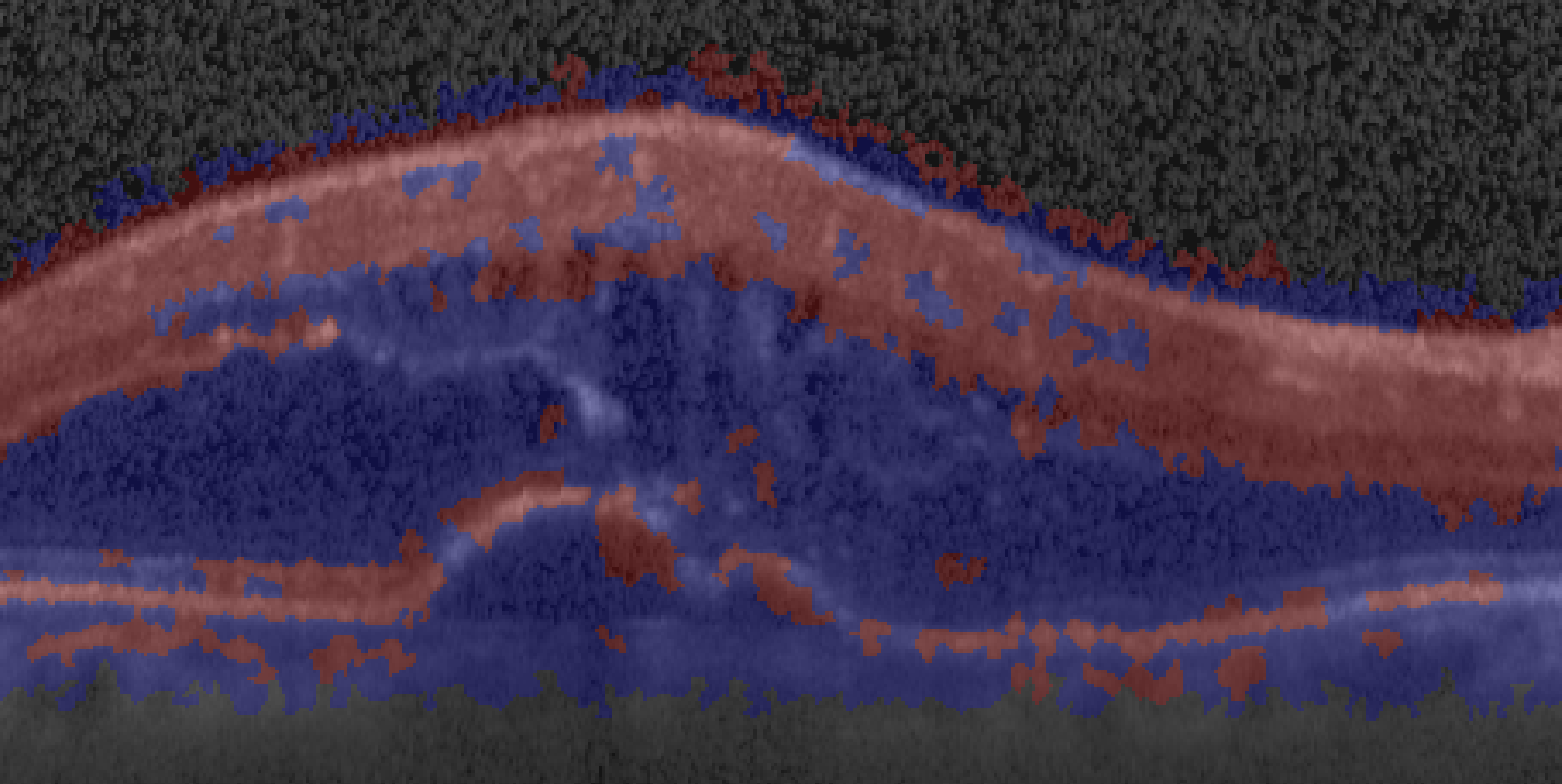}} & 
			\raisebox{-0.5\totalheight}{\includegraphics[height=2cm,width=3.9cm]{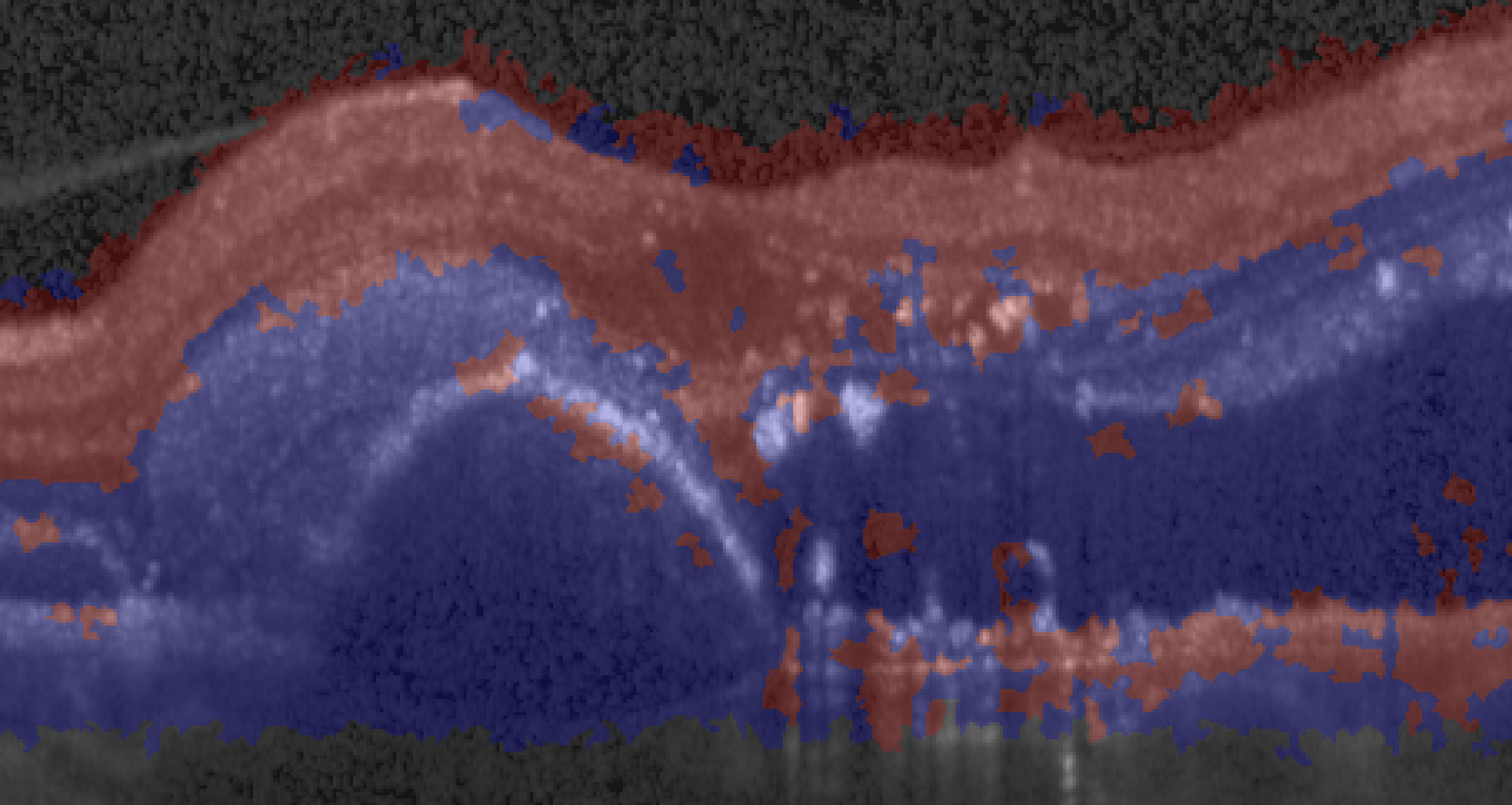}} &
			\raisebox{-0.5\totalheight}{\includegraphics[height=2cm,width=3.9cm]{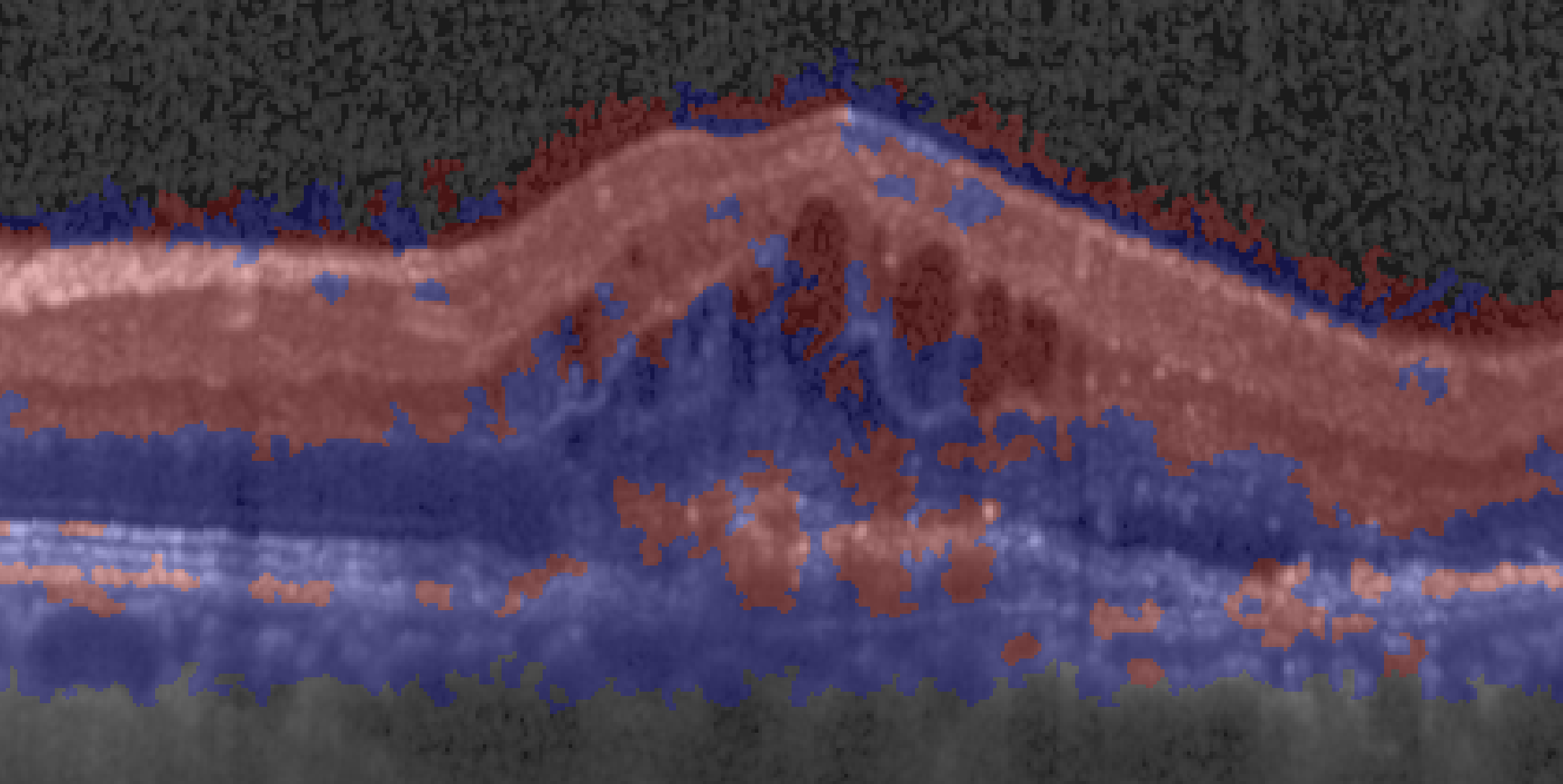}} &
			\raisebox{-0.5\totalheight}{\includegraphics[height=2cm,width=3.9cm]{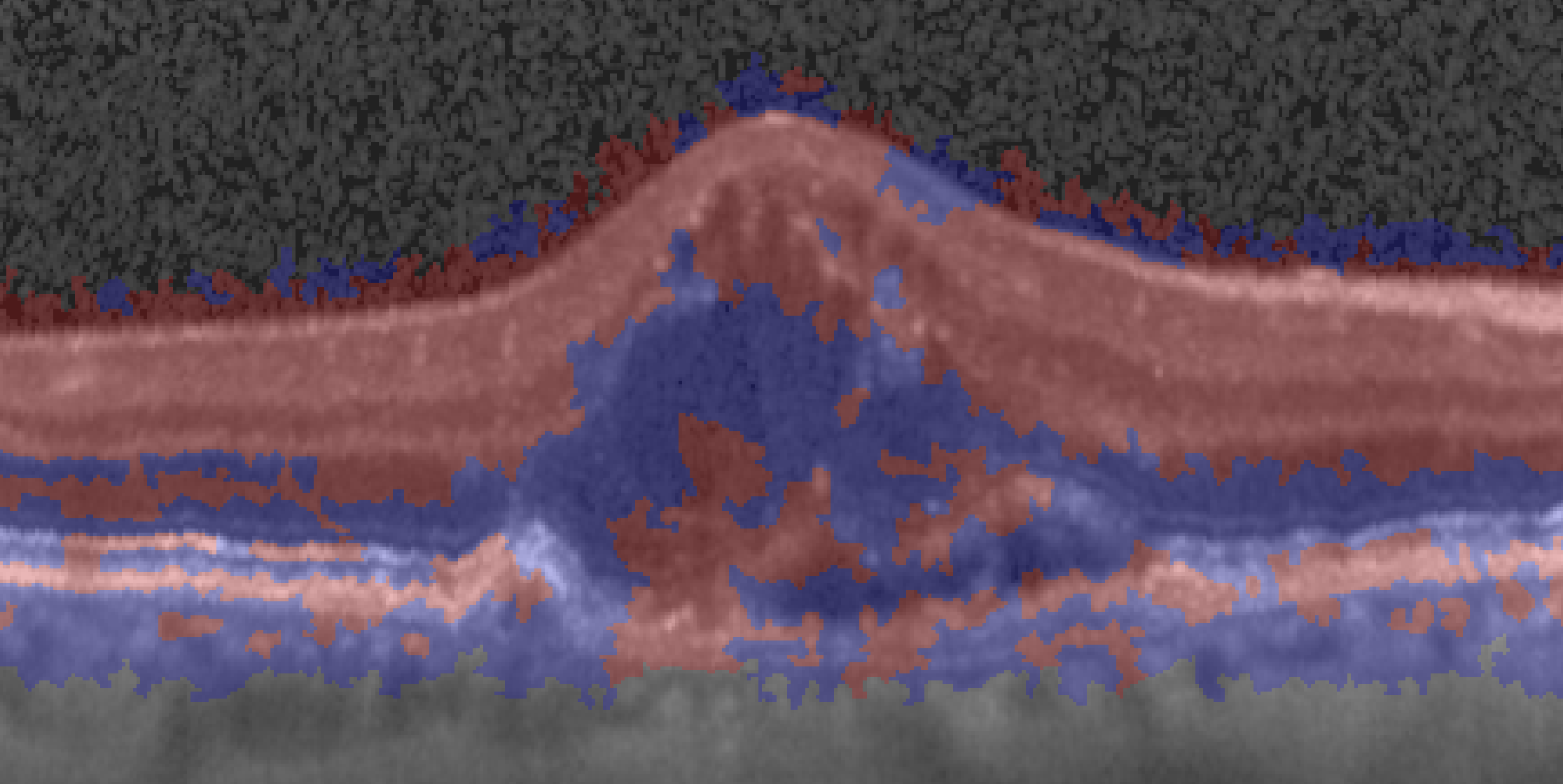}}
			\vspace{3pt} \\
			\makecell[r]{ \scriptsize{$DCAE$} \\ \tiny{\{0.71, 0.67, 0.32, 0.36\}} } &
			\raisebox{-0.5\totalheight}{\includegraphics[height=2cm,width=3.9cm]{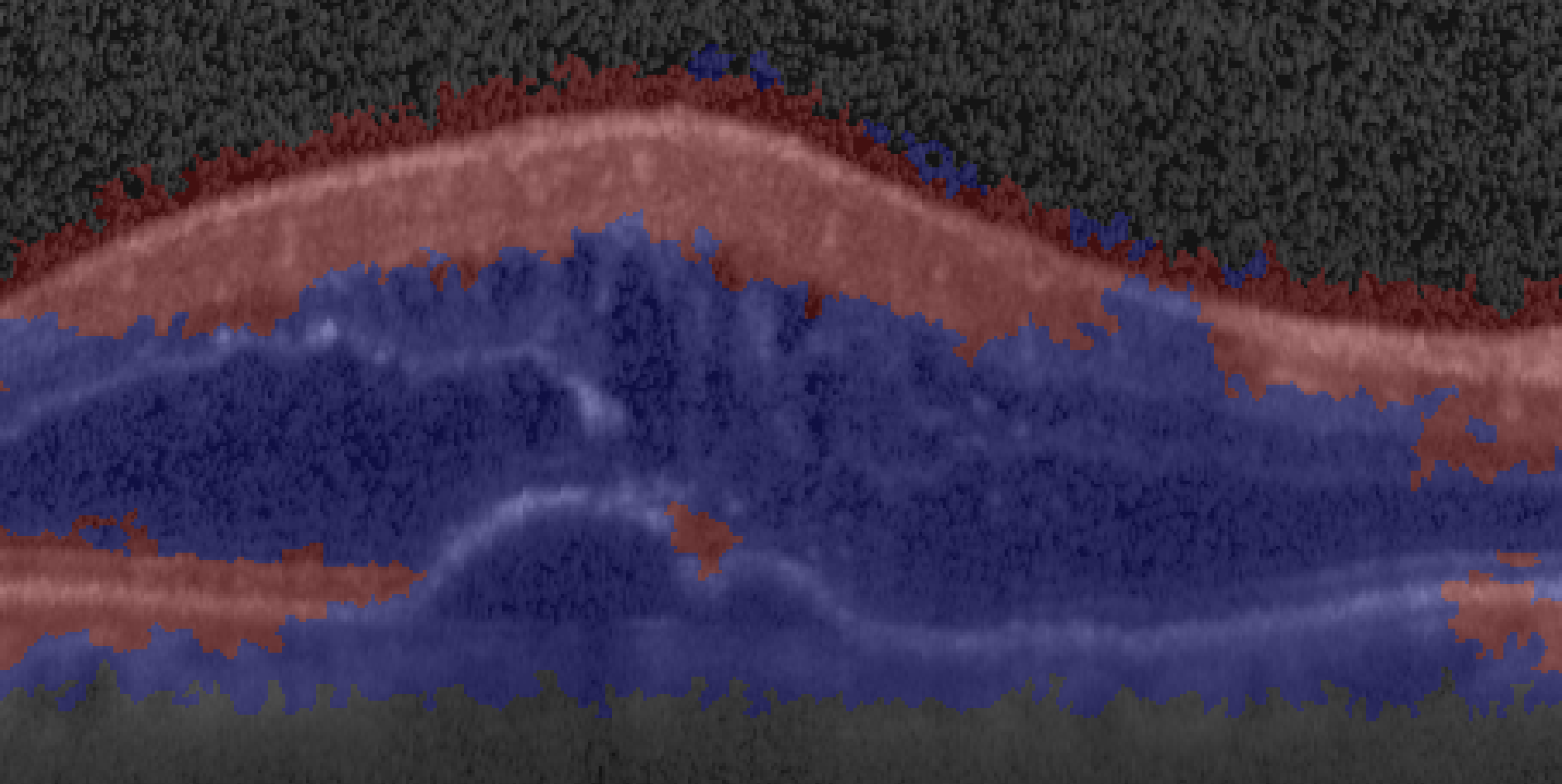}} &
			\raisebox{-0.5\totalheight}{\includegraphics[height=2cm,width=3.9cm]{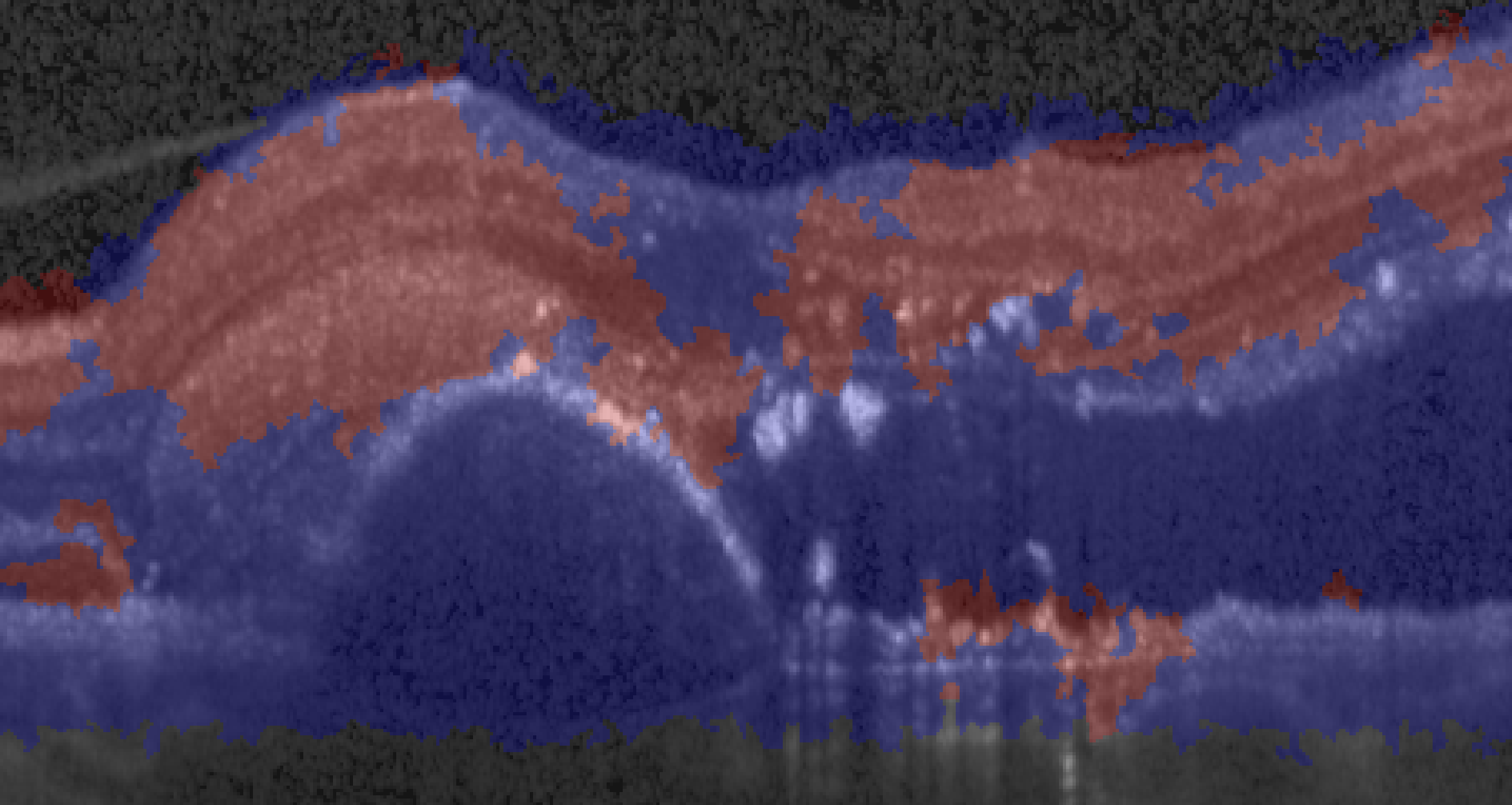}} &
			\raisebox{-0.5\totalheight}{\includegraphics[height=2cm,width=3.9cm]{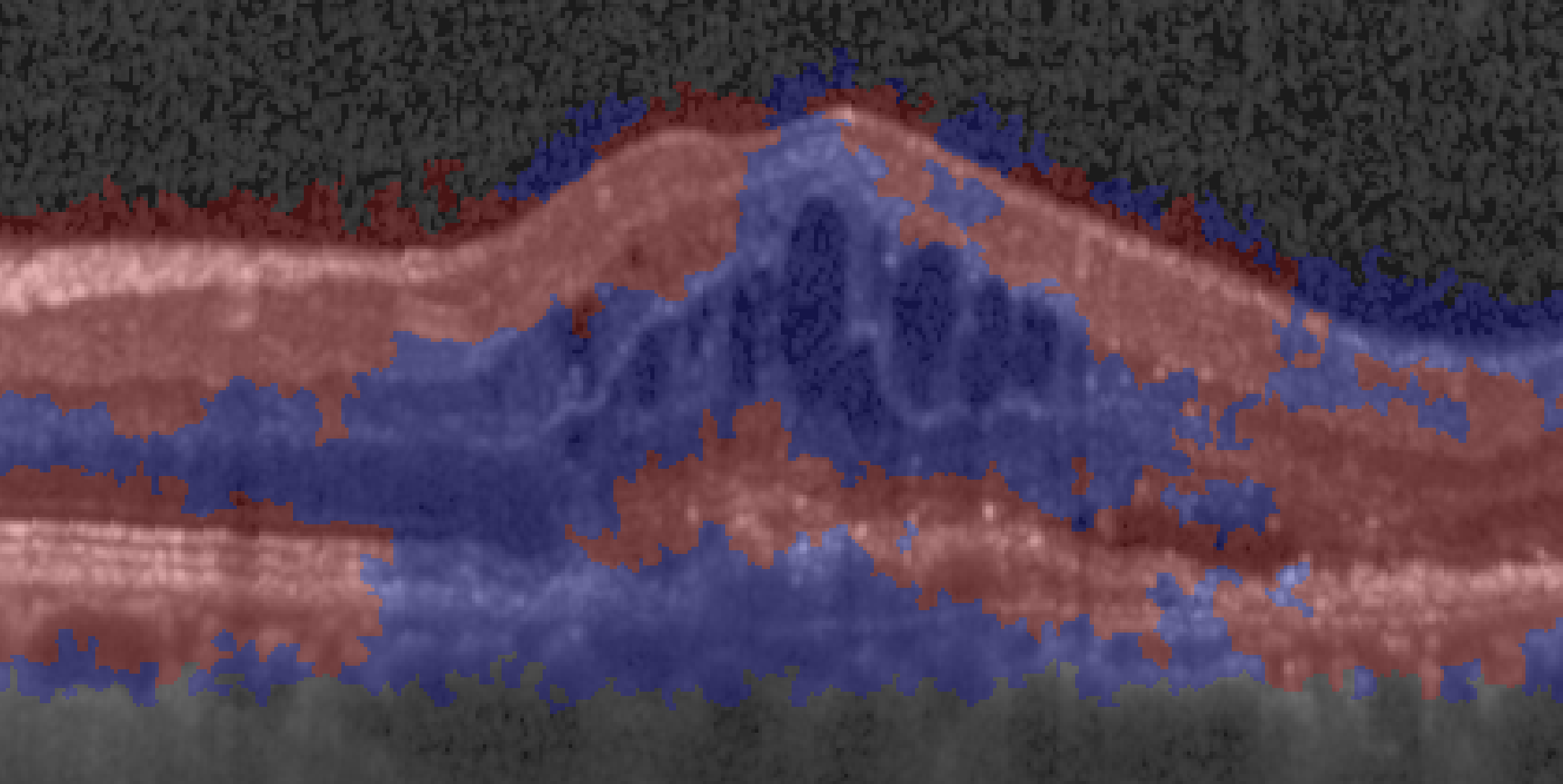}} &
			\raisebox{-0.5\totalheight}{\includegraphics[height=2cm,width=3.9cm]{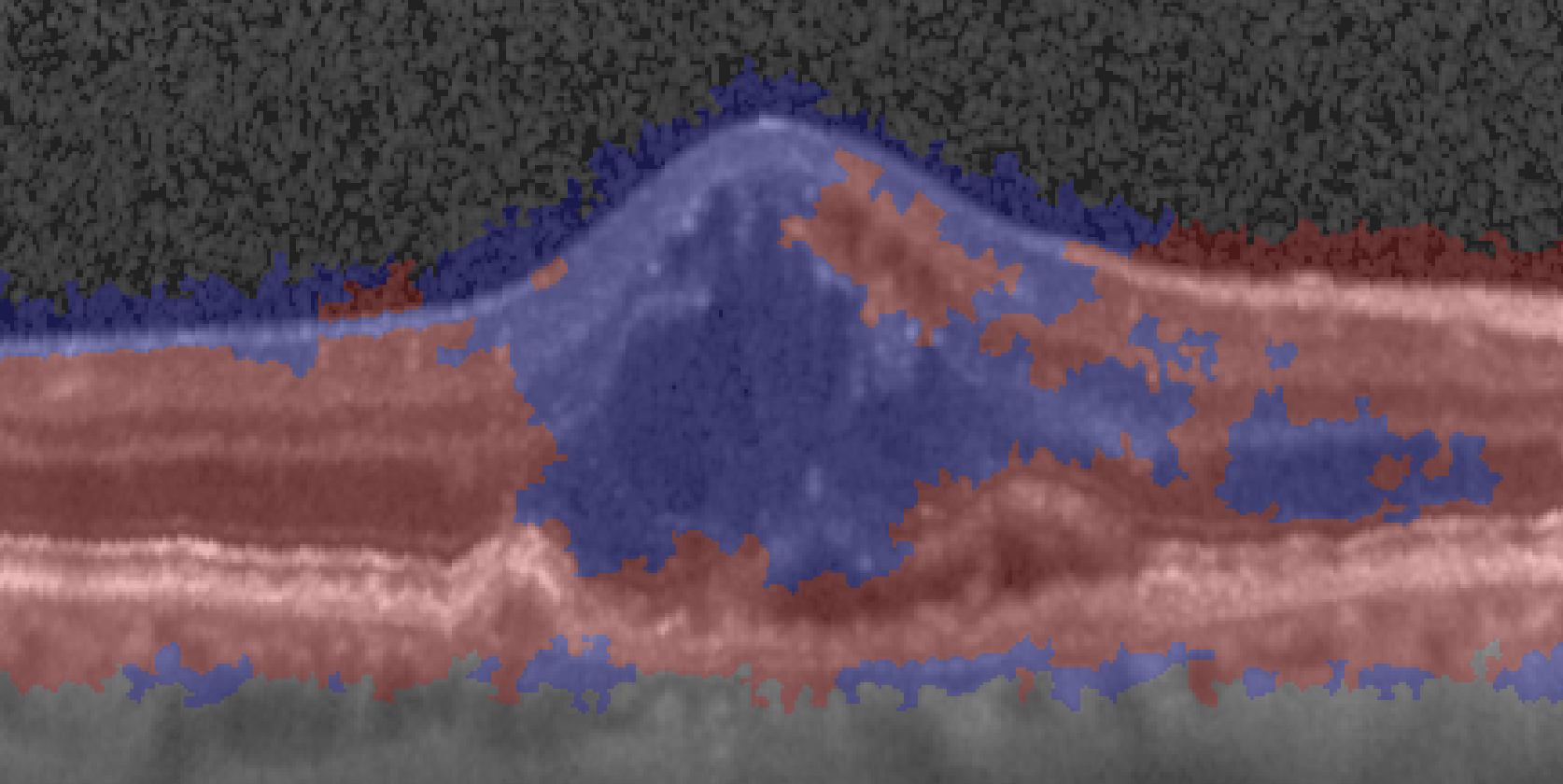}}
			\vspace{3pt} \\
			\makecell[r]{ \scriptsize{$DDAE_{ent}$} \\ \tiny{\{0.68, 0.67, 0.37, 0.35\}} } &
			\raisebox{-0.5\totalheight}{\includegraphics[height=2cm,width=3.9cm]{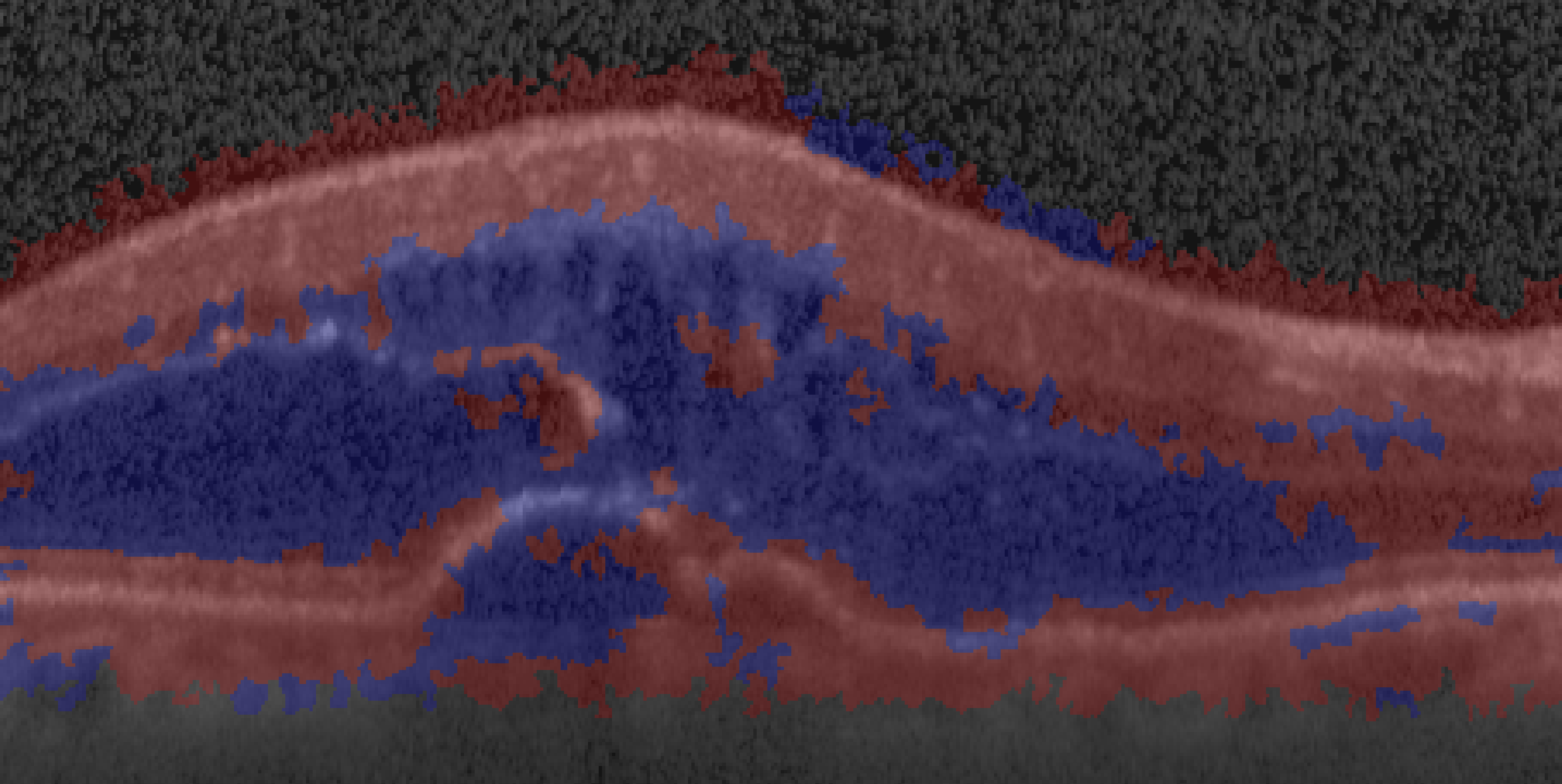}} &
			\raisebox{-0.5\totalheight}{\includegraphics[height=2cm,width=3.9cm]{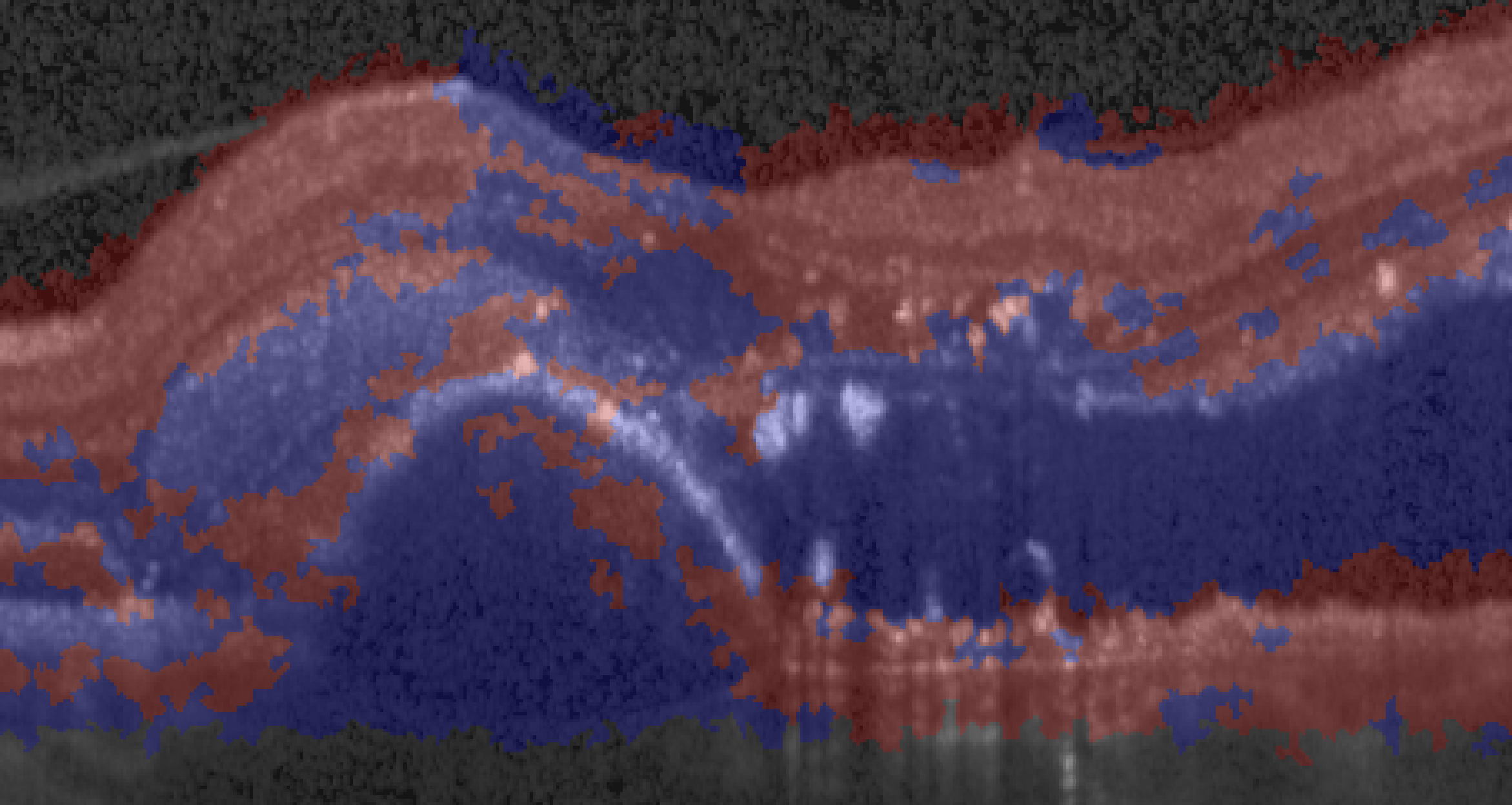}} &
			\raisebox{-0.5\totalheight}{\includegraphics[height=2cm,width=3.9cm]{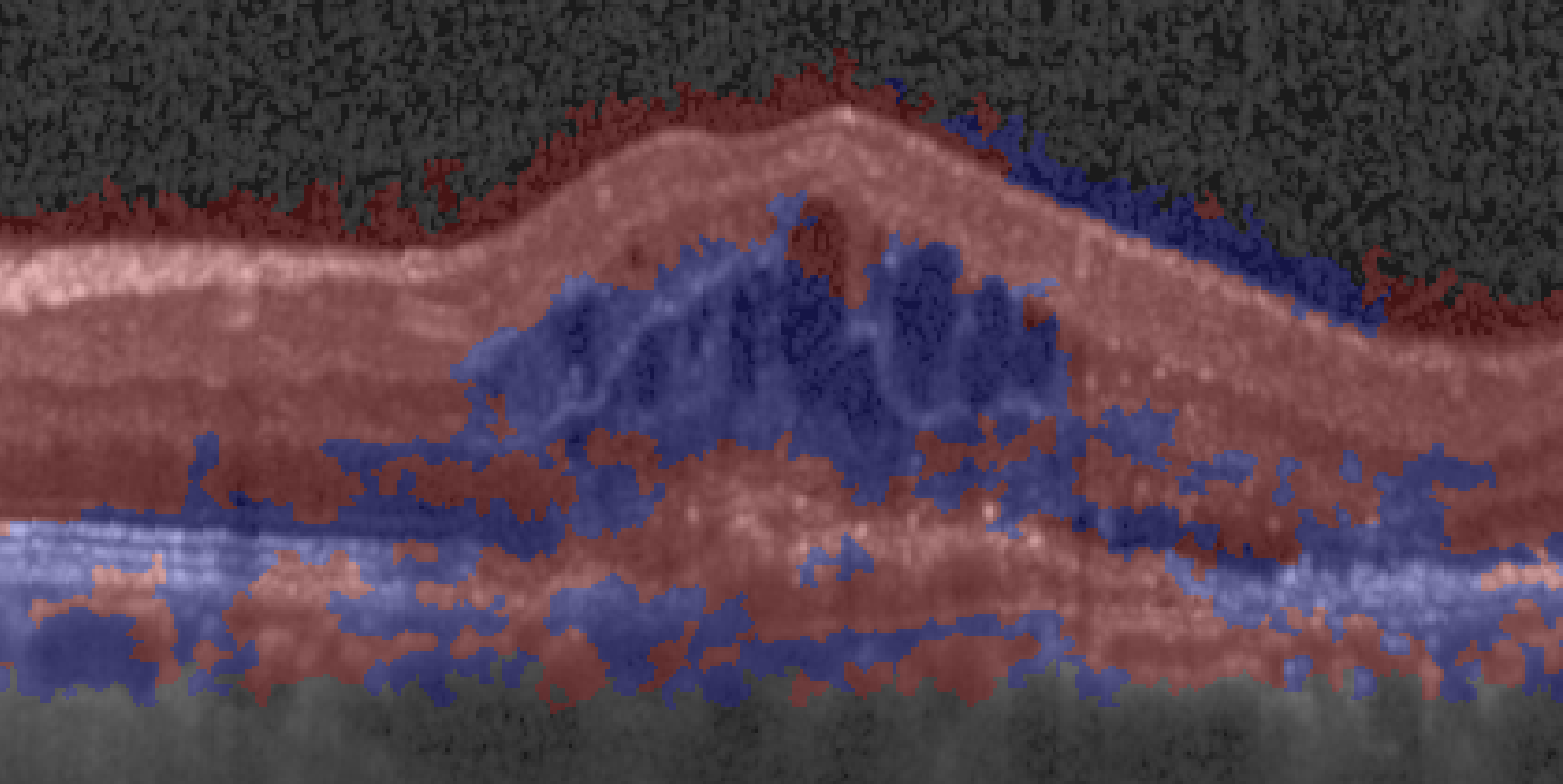}} &
			\raisebox{-0.5\totalheight}{\includegraphics[height=2cm,width=3.9cm]{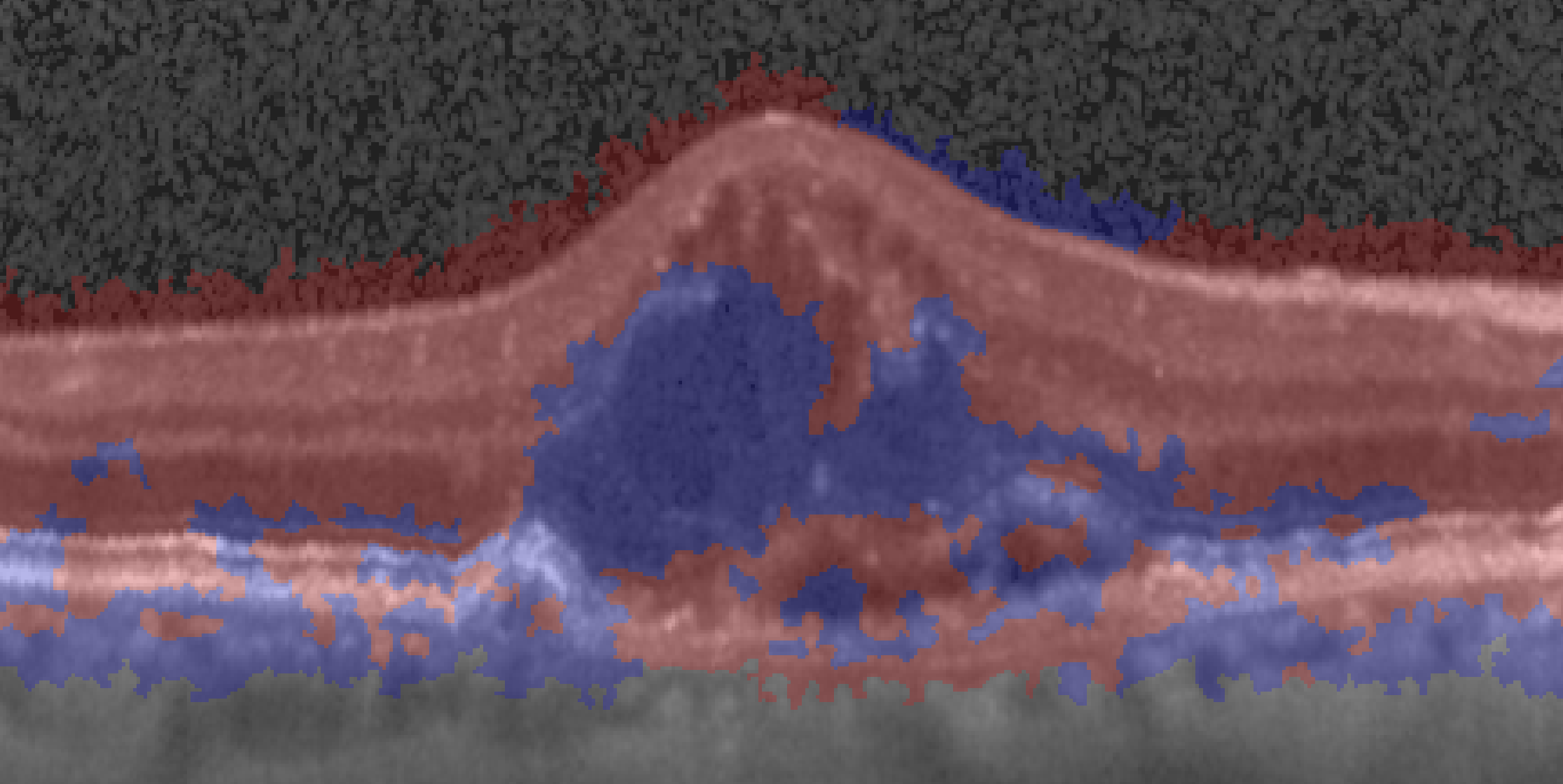}}
			\vspace{0cm} \\
		\end{tabular}
	\end{center}
	\caption{Anomaly detection: for each column, the same B-scan is illustrated, where red and blue colors indicate healthy and anomalous areas. To guarantee objectivity, the first (last) two columns show examples with highest (lowest) dice of $DDAE_{ent}$. Dice values are provided for each method and sample. The full quantitative evaluation result is given in Table~\ref{tab:result_anomaly}.}
	\label{fig:result_anomaly}
\end{figure*}

\paragraph*{Data}
We used n=786 OCT volumes from just as many patients from our database, which was divided into six subsets\footnote{An overview of the data and experiments can be found in the supplementary material}:
\emph{Healthy} (\emph{training} n=283, \emph{test} n=33), \emph{late AMD} (\emph{categorization} n=362, \emph{validation} n=5, \emph{test} n=26), and \emph{early AMD categorization} n=77.
The volumes of \emph{healthy training} and \emph{test} were selected from 482 / 209 contralateral eye scans of patients with RVO / AMD in the other eye. Volumes with pathological changes beyond age-related alterations were excluded.
The volumes of \emph{late AMD} were eyes with active neovascular AMD, where a retina specialist manually annotated all areas that contained pathologic features in 31 OCT volumes. These volumes with voxel-wise annotations of anomalous regions were randomly divided into \emph{late AMD validation} and \emph{late AMD test}.
The volumes in \emph{early AMD categorization} were classified by clinical retina experts as early, non-neovascular AMD according to \cite{ferris2013clinical}.
All image data was anonymized and ethics approval was obtained for the conduct of the study from the ethics committee at the Medical University of Vienna.

The volumes were acquired using Spectralis OCT instruments (Heidelberg Engineering, GER), with a voxel dimensionality of $512\times496\times49$, which depicted a $6mm\times2mm\times6mm$ volume of the retina, with the voxel spacing of  $11{\mu}m\times4{\mu}m\times120{\mu}m$. Thus, one OCT volume is composed of 49 B-scans, where the distance between B-scans is $120{\mu}m$.
All volumes were preprocessed as described in Section \ref{method:preprocess}. Due to the anisotropy of the imaging data, the proposed approach works with 2D patches extracted from B-scans. Pairs of image patches with pixel size of $32\times32$ and $128\times32$ were extracted, illustrated for a single position in Fig.~\ref{fig:method_overallArchitecture}, on the left.

Additionally, we used 384 Bioptigen SD-OCT volumes (269 intermediate AMD, 115 control) from a publicly available dataset~\cite{farsiu2014quantitative}. Since this dataset differs in appearance from our database (different OCT vendor), we conducted additional preprocessing steps: non-local means noise filtering, resizing B-scans to match the Spectralis B-Scans in resolution, and adjustment of image intensity values. Details can be found in the supplementary material.

\paragraph*{Training Details}
All networks were trained on the healthy training set for 300 epochs and used tied weights. A validation set of five OCT volumes was used for parameter tuning. Due to limited computational resources, only a small parameter selection was assessed. We used standard values for ELU ($ \alpha=1 $), momentum (0.9), and mini-batch (50).
The initial learning rate was set to the highest value that did not diverge ($0.0001$) for 150 epochs, and decreased to $0.00001$ for another 150 epochs. We experimented with two different corruption values (0.5, 0.9) for fully connected layers, and we found 0.5 to work better. We also conducted experiments with shallower network architectures, which we empirically found to work slightly worse.
Since the One-Class SVM hyper-parameter $\nu$ is bounded between $0$ and $1$, we varied $\nu$ between $0.01$ and $0.9$ for all methods: $\nu = [0.01, 0.1, 0.2, 0.3, 0.4, 0.5, 0.6, 0.7, 0.8, 0.9]$. 
For the experiments we used the \emph{Torch7} framework~\cite{torch} and the One-Class SVM implementation of \emph{libsvm}~\cite{CC01a}.

\subsection{Evaluation of Anomaly Detection}
\label{evaluation:anomalydetection}
The anomaly detection model was trained on the \emph{healthy training} set, with 277340 extracted pairs of image patches, randomly selected from all $13867$ B-Scans between the top and bottom layer of the retina to avoid retrieving numerous ''background patches'' without relevant content.
While the influence of $\nu$ was analyzed on \emph{late AMD validation} volumes, the final performance of the learned anomaly detection model was evaluated on \emph{late AMD test}. Each B-scan of all validation and test volumes was expert annotated between the ILM and BM layer, and \emph{dice}, \emph{precision}, \emph{recall}, \emph{specificity} and \emph{accuracy} were calculated for algorithmically detected anomalous regions with regard to the manual annotations on volume level.

We compared the $ DDAE_{ent} $ model to the feature learning approaches described in Section~\ref{introduction:relatedwork}: PCA embedding, and a Deep Convolutional Autoencoder, denoted as $ DCAE $. PCA was chosen since it is a well-known and widely used technique for feature learning. At the same time we were aiming at a powerful representation of images, which is why DCAEs, that are specifically designed for images, are a logical comparison method~\cite{zhao2015stacked}.
To ensure a fair comparison for PCA, we trained two models. In the first model, $PCA_{256}$, the dimensionality was chosen to match the feature dimension, $z$, of the proposed model. For both scales, the first 128 principal components were kept. In the second model, $PCA_{0.95}$, for each scale, the first components that described 95\% of the variance were kept. To retrieve the final feature representation, the feature vectors $ \dot{y}$ and $\ddot{y}$ of both scales were concatenated to obtain $ z $.

For $DCAE$, we use an encoder with a convolutional layer (\texttt{c}) and 512 $9\times9$ filters, followed by a $3\times3$ non-overlapping max pooling (\texttt{p}) and two fully connected layers with 2048 and 512 units (\texttt{512c9-3p-2048f-512f}). The decoder was composed of deconvolution (\texttt{dc}) and unpooling (\texttt{up}) layers to approximately invert the output of the encoder and reproduce the input (\texttt{512f-2048f-3up-512dc9}). All layers, except pooling and unpooling, were followed by ELUs.
This DCAE-architecture replaced $ DDAE_{1} $ and $ DDAE_{2} $, while the architecture of the third model($ DDAE_{3} $) remained the same. To ensure a fair comparison, the feature dimension of the individual model outputs matched the dimensionality of the proposed method.

\subsection{Evaluation of Anomaly Categorization}
\label{evaluation:anomalyCategorization}
We trained two clustering models on two different datasets: \emph{late AMD categorization} and \emph{early AMD categorization}. We extracted 354760 and 75460 pairs of image patches, respectively.
For both models, we varied the number of clusters, $C$, between 2 and 30 and selected the clustering model with lowest DB-index. 
In order to qualitatively evaluate the categories found in the regions identified as anomalous, a segmentation of the retina based on the identified anomalous categories was computed on both datasets. Assignment of each pixel was based on the learned centroids and the nearest-cluster-center labeling (Section \ref{method:clustering}).
The used manual annotations of the test set provided only a binary distinction into healthy and anomalous, and did not describe all anomalies that were visible in separate categories. Therefore, two clinical retina experts conducted a qualitative evaluation of the results by visually inspecting the results. While the number of clusters was determined by the DB-index, category descriptions were identified by the experts. Additionally, cosine distances between centroids of the two cluster models trained on \emph{late AMD categorization} and \emph{early AMD categorization} were computed in order to evaluate the correspondence between both models.

\subsection{Evaluation of Volume Level Disease Classification}
\label{evaluation:volumeclassification}
To evaluate if the identified categories can serve as disease markers and encode valuable discriminative information, we used the segmentation of the retina into $C$ clusters, originating from the clustering model learned on \emph{late AMD categorization}, to conduct multi-class classification on patient level. The volume of each cluster served as feature vector for every case. Since the clusters that were identified in \emph{early AMD categorization} could all be mapped to clusters identified in \emph{late AMD categorization}, which at the same time revealed one additional cluster, we used only the latter more complete anomaly category set as basis for these experiments.
We trained a random forest (RF) classifier \cite{breiman2001random} (\#trees=64) with these feature vectors using a set of randomly chosen late AMD, early AMD and healthy cases (n=50 per class) from the training and categorization sets.
We then applied the classifier to a separate test set not involved in anomaly detection, categorization, or classifier training composed of \emph{late AMD test} (n=26) the remaining part of \emph{early AMD categorization} (n=27) and \emph{healthy test} (n=33). We report feature importance values obtained by random forest training, and the classification accuracy on the test set.

For comparison, we trained a second RF model without category information. Using the same evaluation setting as described above, we used the binary segmentation of the retina as features, originating from anomaly detection (Section \ref{method:anomaly}), instead of the learned clusters.

We performed a second experiment to evaluate how the method generalizes to a dataset of a different vendor. The Bioptigen volumes were used for a second volume classification experiment. Following the evaluation procedure in~\cite{venhuizen2015automated}, the dataset was randomly divided into \emph{bioptigen training}~(218 AMD, 65 control) and \emph{bioptigen test}~(50 AMD, 50 control), and the RF was trained with \#trees=100. Again, we trained two models with and without category information, originating from the clustering model learned on \emph{late AMD categorization}.

\section{Results}
\label{results}
We report quantitative and qualitative results that illustrate anomaly detection, visualize anomaly categorization outcome, provide descriptions of clusters according to experts and describe results of volume disease classification tasks using the identified categories as marker candidates.

\subsection{Anomaly Detection Results}
\label{results:anomaly}

\begin{table*}[t]
	\begin{minipage}[t][][b]{0.65\linewidth}
		\centering
		\caption{Dice, precision, recall (=sensitivity), specificity, and accuracy for anomalous regions with manual annotations, calculated on the test set. Additionally, the chosen $\nu$ value for the One-Class SVM is reported.}
		\label{tab:result_anomaly}
		\begin{tabular}{l|lllll}
			\toprule
			Algorithm ($\nu$) & Dice & Precision & Recall & Specificity & Accuracy \\
			\midrule
			$PCA_{256}$ \textit{(0.4)} &  0.47 (0.12) & 0.36 (0.12) & 0.74 (0.09) & 0.46 (0.03) & 0.55 (0.05) \\ 
			$PCA_{0.95}$ \textit{(0.2)}  &  0.51 (0.12) & 0.40 (0.13) &  0.74 (0.08) & 0.56 (0.04) & 0.62 (0.04) \\ 
			$DCAE $ \textit{(0.2)} &  0.49 (0.13) & 0.41 (0.13)  & 0.64 (0.14) & 0.63 (0.07) & 0.65 (0.05) \\ 
			$\bf{DDAE_{ent}}$ \textbf{\textit{(0.1)} }& \textbf{0.53 (}\textbf{0.09)} & \textbf{0.47 (}\textbf{0.12)} & \textbf{0.63 (}\textbf{0.06)} & \textbf{0.71 (}\textbf{0.07)} & \textbf{0.69 (}\textbf{0.05)} \\
			\bottomrule
		\end{tabular}
	\end{minipage}
	\hfill
	\begin{minipage}[t][][b]{0.30\linewidth}
			\centering
			\includegraphics[width=0.9\textwidth]{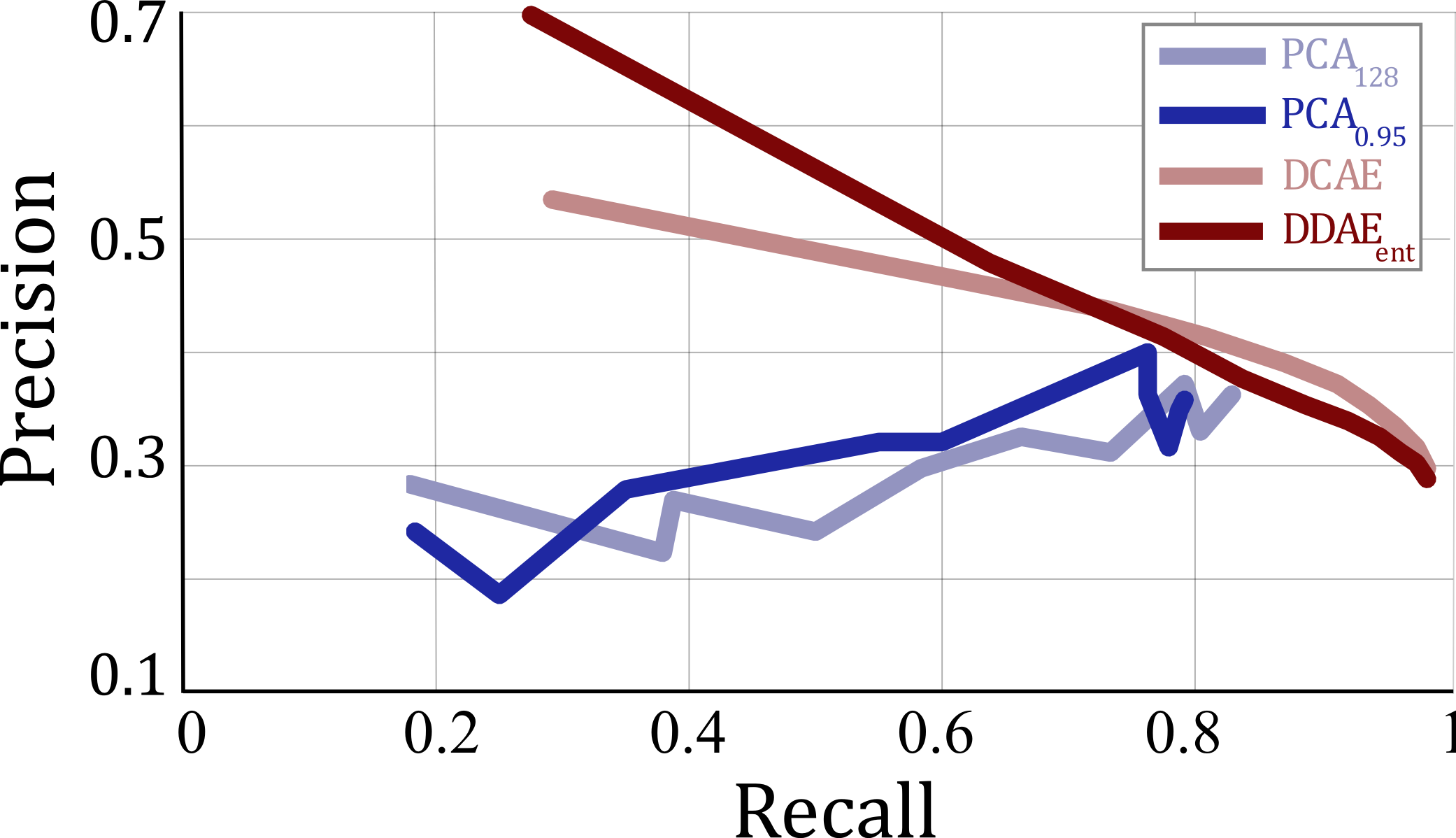}
			\captionof{figure}{precision-recall curve, calculated on the validation set.}
			\label{fig:result_PrecRecCurve}
	\end{minipage}
\end{table*}

For the detection and segmentation of anomalies, the proposed method achieved a dice of $0.53$ between annotated and predicted anomalous regions, a precision of $0.47$, and a recall of $0.63$, which means that 63\% of all manually annotated anomalies were also identified as anomalous by our model (Table~\ref{tab:result_anomaly}). $PCA_{256}$, $PCA_{0.95}$ and $DCAE$ achieved a lower Dice ($0.47$, $0.51$, and $0.49$) compared to our method. Using a \emph{paired Wilcoxon signed-rank test}, a significant difference could be shown for $PCA_{256}$~(p=0.0004) and $DCAE$~(p=0.02), but not for $PCA_{0.95}$~(p=0.11).

To enable an objective qualitative evaluation, the volumes which are visualized in Fig.~\ref{fig:result_anomaly} were selected according to highest and lowest dice of  $DDAE_{ent}$.
An additional visual comparison of the segmentation results revealed that the shape of identified anomalous regions of the proposed method, $DDAE_{ent}$, reflected the manual annotations better than all comparison methods.

The validation performance for all examined $\nu$ values and all methods is reported in Fig.~\ref{fig:result_PrecRecCurve} and Fig.~\ref{fig:result_nuVSprecisionRecall}.
At a recall level around 0.78, where the precision-recall curve (Fig.~\ref{fig:result_PrecRecCurve}) seems to reveal comparable performance of the examined methods, $DDAE_{ent}$ achieves a precision of $0.42$, outperforming all other approaches.
At the same time, when comparing the curves, it can be clearly observed that both $DCAE$ and $DDAE_{ent}$ produced more stable results in comparison with the PCA methods.
In particular, Fig.~\ref{fig:result_nuVSprecisionRecall} shows that precision/recall decreased/increased continuously as ν increased for $DCAE$ and $DDAE_{ent}$, while both PCA methods exhibited an inconsistent behavior. In accordance with these quantitative outcome, Fig.~\ref{fig:result_consistency} illustrates segmentation results for $DDAE_{ent}$ and $PCA_{0.95}$. Note that the embedding itself did not change with varying  $\nu$ values. This inconsistency of both PCA methods makes an intuitive interpretation and adaption of $\nu$ difficult, though it may be important for specific tasks to control the precision-recall trade off.

\begin{table*}[t]
	\begin{minipage}[t][][b]{0.65\linewidth}
	\centering
	\includegraphics[width=\textwidth]{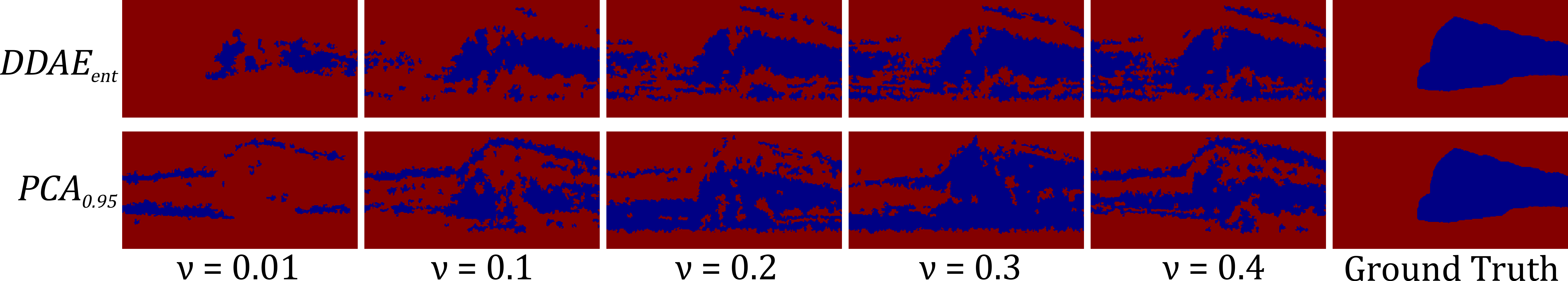}
	\captionof{figure}{Compared to the PCA methods, $DDAE_{ent}$ produced more stable results when varying $\nu$. This finding was also supported by the segmentation results, illustrated for five consecutive $\nu$ values of $DDAE_{ent}$ and $PCA_{0.95}$ on an example B-Scan (anomaly regions are highlighted in blue).}
	\label{fig:result_consistency}
	\end{minipage}
	\hfill
	\begin{minipage}[t][][b]{0.30\linewidth}
		\centering
		\includegraphics[width=0.84\textwidth]{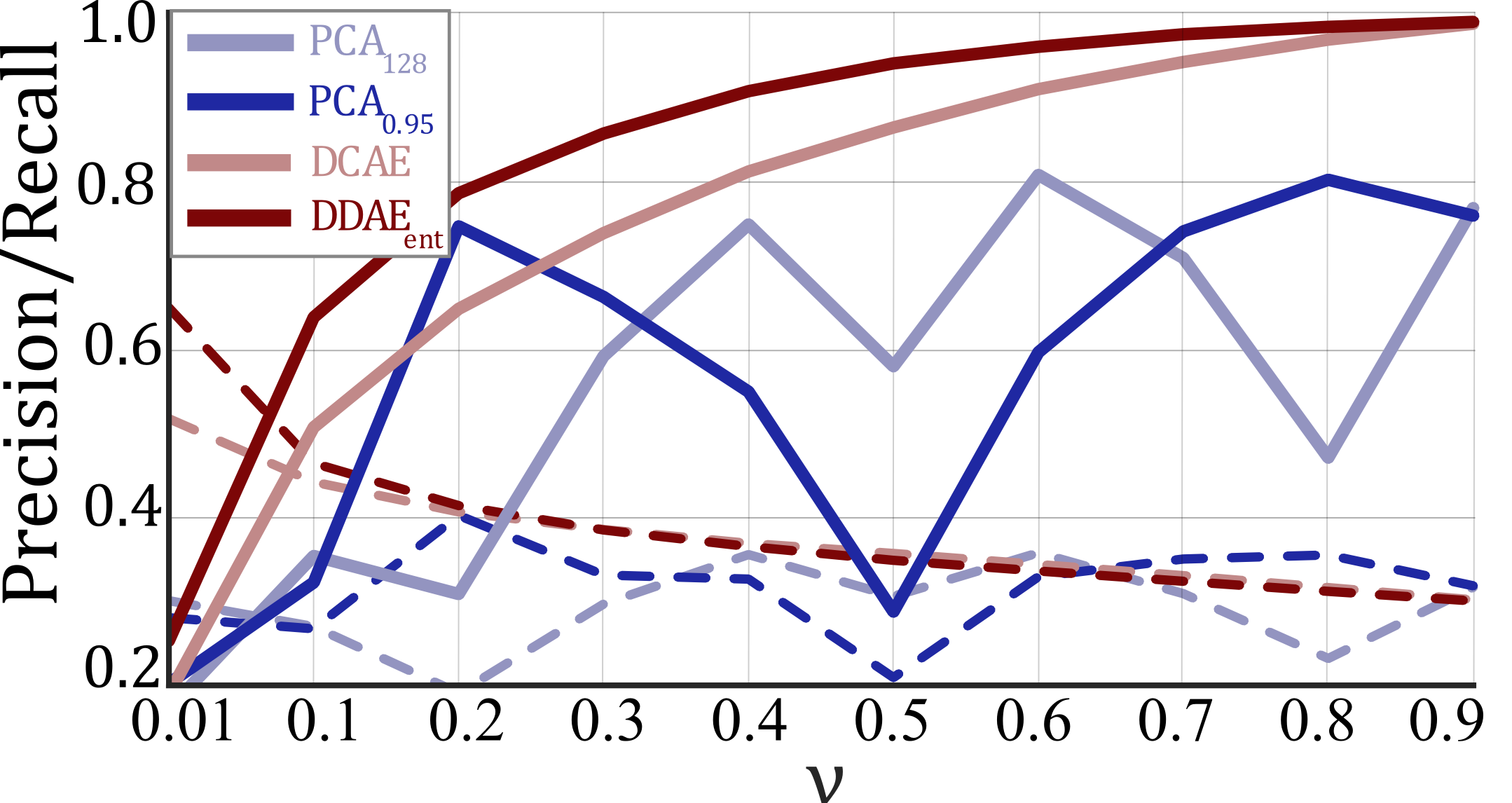}
		\captionof{figure}{Validation values of precision (dashed) and recall (solid) are plotted against $\nu$.}
		\label{fig:result_nuVSprecisionRecall}
	\end{minipage}
\end{table*}

\subsection{Anomaly Categorization Results}
\label{results:clustering}
\begin{figure*}[t]
	\centering
	\includegraphics[height=17cm,clip=true,trim=0.5cm 7cm 4cm 0cm]{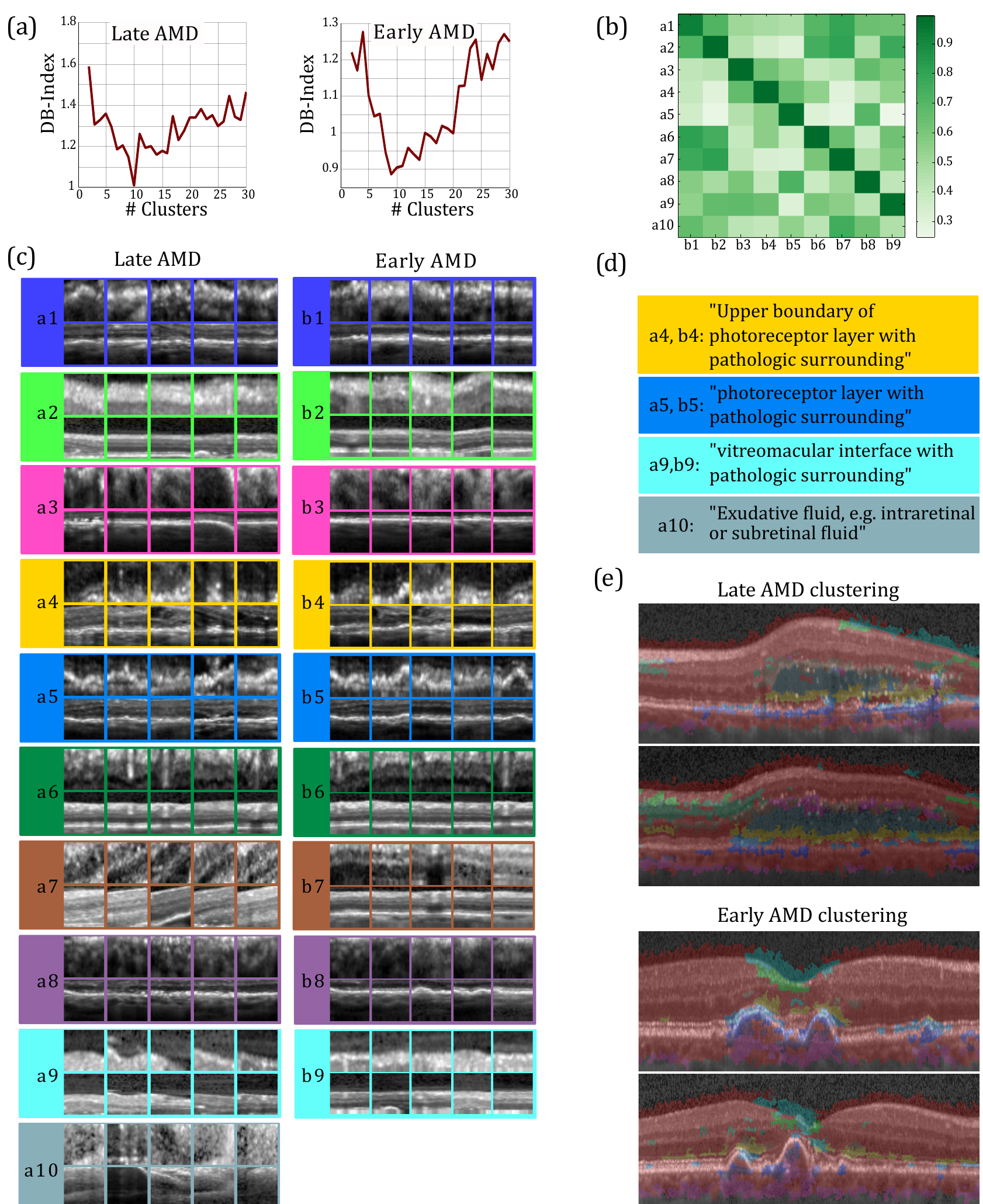}
	\caption{Anomaly categorization: The calculated values of the DB-Index are plotted in (a). The cosine distance between cluster centroids is visualized in (b) and is bounded between 0 and 1, due to rescaling of features  into positive domain. The nearest-neighbors of cluster centroids are illustrated in (c). The upper row shows the 32-by-32 patches, while the second row illustrates the 124-by-32 patches. Each cluster is indicated by a separate color. Some exemplary cluster descriptions that were identified by experts are denoted in (d).
	Clustering results of \emph{late AMD} and \emph{early AMD clustering} are shown in (e) on example B-scans, where identified anomalous regions were segmented into 10 and 9 categories, respectively. In accordance with former visualizations, normal regions are highlighted in red.}
	\label{fig:result_clusteringOCT}
\end{figure*}

Despite the fact that the anomaly detection performance left room for improvement in general, the detected anomaly candidates could be clustered into stable categories. The lowest DB-index was found for $C=10$ on \emph{late AMD categorization} and $C=9$ on \emph{early AMD categorization}, as indicated in Fig.~\ref{fig:result_clusteringOCT}(a). This was a plausible outcome, since OCT volumes with late AMD exhibit more obvious visual variation than early AMD volumes.

The cosine distance between cluster centroids is visualized in Fig.~\ref{fig:result_clusteringOCT}(b), where the columns were re-arranged for better interpretability. The nearest-neighbors of cluster centroids are illustrated in Fig.~\ref{fig:result_clusteringOCT}(c), both for \emph{late AMD} and \emph{early AMD clustering} results.
As can be seen both in \ref{fig:result_clusteringOCT}(b) and (c), all clusters of \emph{early AMD clustering} could be linked to specific clusters in \emph{late AMD clustering}. This was a plausible outcome, since all variation that is present in \emph{early AMD}, is also present in \emph{late AMD} cases. 
Exemplary category descriptions identified by experts are denoted in Fig.~\ref{fig:result_clusteringOCT}(d), where ''Upper boundary of photoreceptor layer with pathologic surrounding'' (a4, b4), ''Photoreceptor layer with pathologic surrounding'' (a5, b5), and ''vitreomacular interface with pathologic surrounding'' (a9, b9) could be identified in both clusterings.

In contrast, \emph{late AMD clustering} showed one additional cluster ''a10'' which was identified as ''Exudative fluid" (e.g. intraretinal or subretinal fluid) segmentation by the clinical retina experts, and had no clear relation to a specific \emph{early AMD} cluster. This claim of missing relation was supported by qualitative evaluation as well as by the calculated cosine distance between cluster centroids, which showed relatively low values (large distances) for ''a10'' to all \emph{early AMD} clusters, as illustrated in Fig.~\ref{fig:result_clusteringOCT}(b), bottom row. Clustering results are shown in Fig.~\ref{fig:result_clusteringOCT} (e) on \emph{late AMD test} B-scans, where it can be seen that cluster ''a10'' showed substantial overlap with areas of fluid.
Since fluid like intra- or sub-retinal fluid occurs only in late AMD, this was a reasonable outcome and indicated that also disease specific clusters had been learned.

\subsection{Volume Level Disease Classification Results}
\label{results:volumeclassification}
We obtained an accuracy of 81.40\% on the three-class classification task, using the volume of each cluster (corresponding to \emph{late AMD clustering}) as features. The confusion matrix (Fig.~\ref{fig:result_featImp}(a)) shows that the classifier could successfully distinguish between late and early AMD cases. It is a more difficult task to separate early AMD and healthy volumes\footnote{Result examples can be found in the supplementary material.}. The feature importance, calculated during  random forest training, is given in Fig.~\ref{fig:result_featImp}(b). It visualizes how each feature contributes to the prediction of a class in the form of the mean decrease of accuracy (MDA) for individual feature perturbations.
We provide information about whether variables are positive or negative predictors by comparing their average value within class examples to the average value for out-of-class examples as the sign. Results identify ''a7'' as the most important feature of the calculated random forest model. It is a strong negative predictor for healthy, while a strong positive predictor for late AMD. The comparison experiment without category information resulted in lower accuracy of 60.47\% on the same classification task. 

On the \emph{bioptigen test} set, we achieved an area under the ROC curve (AUC) of 0.944 (Figure~\ref{fig:result_aucCurve}) and 0.768 for the RF models using category information or not, respectively. The clear performance gaps on both classification tasks supports the claim that the learned anomaly categories encode clinically meaningful information. Furthermore, the result on \emph{bioptigen test} indicates that the learned feature representation and the categories, respectively, reflect morphological properties of the retina, and not OCT vendor specific characteristics.

\begin{table*}[t]
	\begin{minipage}[t]{0.70\linewidth}
		\centering
		\includegraphics[height=3.8cm]{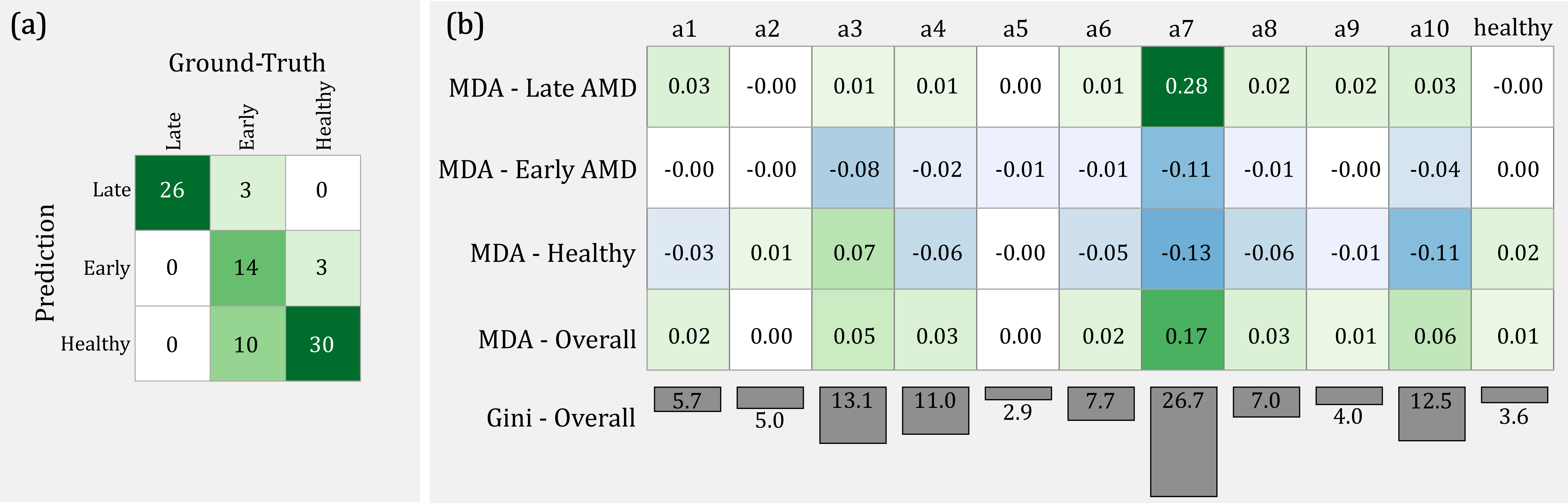}
		\captionof{figure}{Three-Class Classification: (a) Confusion matrix of the test set. (b) Identified anomaly categories as markers of disease. The first three rows show the class-specific MDA, where the sign encodes the feature-trend for that specific class (positive indicates high within class and low outside class values and vice versa). The fourth row contains the MDA over all classes, and the last row shows the mean decrease in Gini index.}
		\label{fig:result_featImp}
	\end{minipage}
	\hspace{0.2cm}
	\begin{minipage}[t]{0.25\linewidth}
		\centering
		\includegraphics[height=3.8cm]{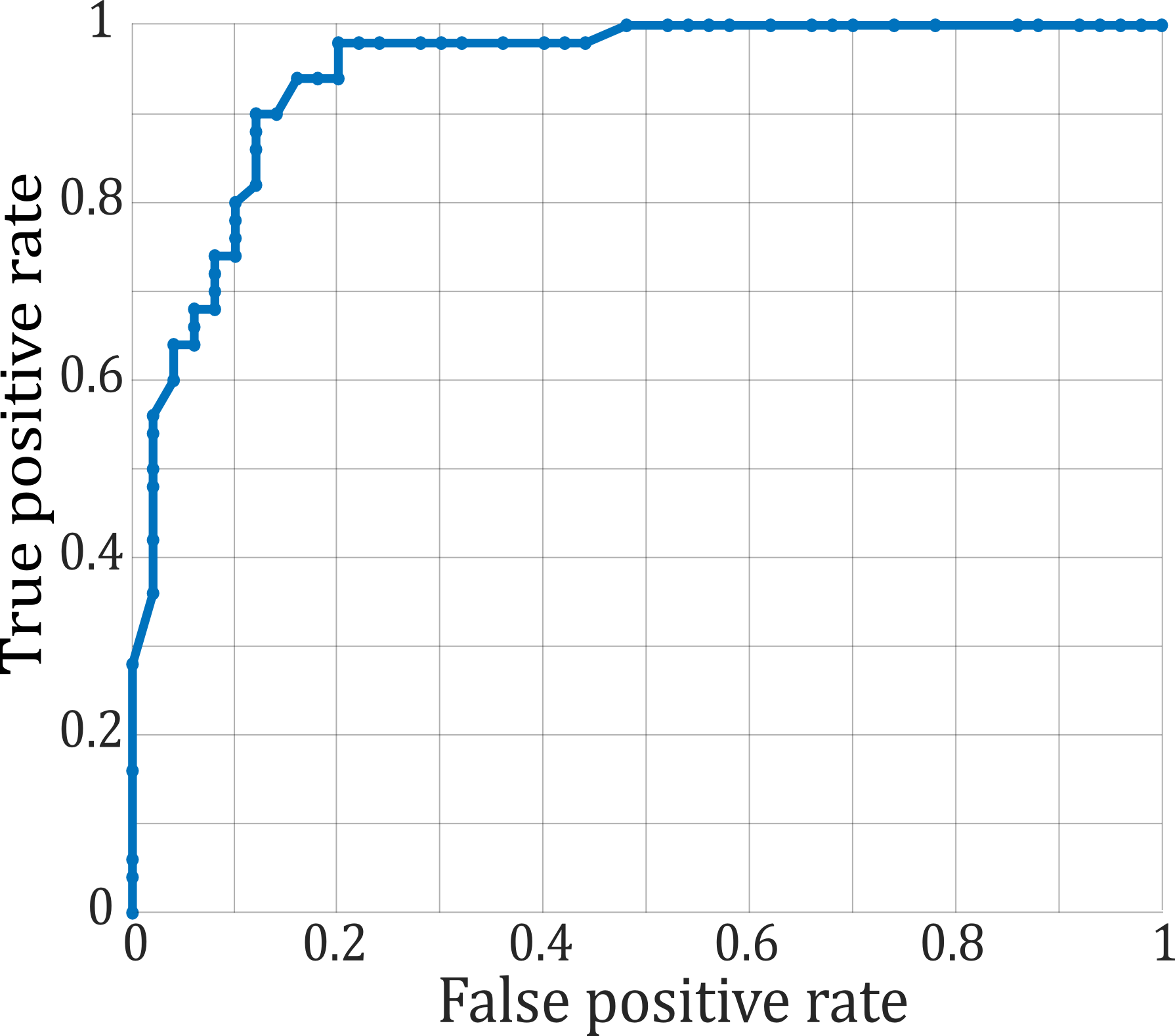}
		\captionof{figure}{ROC curve of the Bioptigen binary classification task.}
		\label{fig:result_aucCurve}
	\end{minipage}
\end{table*}

\section{Discussion}
We propose a method to detect and categorize anomalous regions in OCT volumes of the retina, and subsequently use these anomalies as marker candidates. The model is trained on healthy imaging data and detects anomalies in new volumes without constraints to a priori definitions. Categorization of  anomalies revealed clusters of frequently occurring patterns, where a part of these categories could be mapped to clinically meaningful entities in the imaging data in a post hoc qualitative assessment of clusters by experts.
Finally, results in disease classification tasks indicate that the identified marker candidates encode valuable discriminative information.

\paragraph{Three insights}
From evaluation results we gain three primary insights. First, the proposed approach relying on a multi-scale deep denoising auto encoder architecture to represent image information shows comparable or superior performance to alternatives such as PCA or DCAE. At the same time, the embedding of $DDAE_{ent}$ allows to control the precision-recall trade off in an intuitive way, as opposed to PCA. This indicates that the representation is important for successful detection of subtle alterations in the imaging data and stable training of the one-class SVM.

Second, we can identify stable categories, that are replicable across data sets. Clustering reveals entities that are present in late- and early AMD, and a class of entities that is only present in late AMD. It demonstrates that purely data driven learning can reveal meaningful structure in the data, that corresponds to disease processes. Here, it reflects the emergence of exudative liquid that is characteristic for late AMD.

Third, the identified anomaly categories are valid marker candidates, that show predictive value, when used for volume level classification.

\paragraph{Relationship to prior work}
While we achieved an AUC of 0.944 on the binary classification task, prior work reported an AUC of 0.984~\cite{venhuizen2015automated} and 0.992~\cite{farsiu2014quantitative} on the Bioptigen dataset, where the latter used a different evaluation process (leave-one-out cross-validation on all cases).
In~\cite{venhuizen2015automated} features are extracted at interest points which are located using manually defined constraints, while Farsiu \etal~\cite{farsiu2014quantitative} used semi-automatically segmented retinal layers as features. In contrast to our study, both works use prior knowledge about the disease to create features specifically designed for this classification task.
Additionally, our features are generated by a model which was trained on cases from a different OCT device (Spectralis vs. Bioptigen), which adds additional complexity to the task. Viewed in this light, our result indicates that the learned anomaly categories encode valuable discriminative information.

\paragraph{Identification of novel marker candidates}
There is strong interest in the identification of valid biomarkers in AMD, since the already known biomarkers (e.g., retinal thickness, macular fluid) do not explain the entire spectrum of the disease and in particular the individual level of vision loss~\cite{SchmidtErfurth20161}. The proposed method contributes a path to identify novel marker \emph{candidates}. It found categories that were known (e.g. photoreceptor layer with pathologic surrounding, cluster a5/b5), as well as potential new biomarker categories such as ''a7'', which could not clearly be linked to a particular known pathology by the clinical retina experts and at the same time showed a high feature importance regarding disease classification.
The ultimate aim is to use unsupervised automated analysis to identify disease marker candidates in a first step, as done in this study, and to define a precise description of characteristics of those candidates in a second step, transforming them from candidates to effective markers applicable in clinical practice. The latter is subject of future work, for instance by correlating marker candidates with visual function.
Results showing that the identified categories can classify disease are the strongest indication that unsupervised learning as proposed in this paper, can identify novel marker candidates and potentially contribute to understanding mechanisms governing disease course and treatment effect.
If accuracy of the method can be improved further, in addition to marker identification, future work could also use anomaly detection to quickly visualize anomalies in OCT volumes, helping to efficiently evaluate large datasets, or in a screening setting.

\paragraph{Limitations}
There are some limitations that have to be mentioned.
First, the performance of the pixel-wise anomaly detection (dice=0.53) left room for improvement. While a recall of 0.63 indicates that manually annotated regions were still missed in this step, the relatively low precision of 0.47 may result from two sources: First, normal appearance dissimilar to the range represented in the training set, due to not having enough training data. The second possible source may be anomalies that have not yet been categorized, and are potential new candidates for markers.
The interpretation of identified marker candidates remains challenging. They do not correspond to known categories, and thus no ground-truth exists for their direct evaluation. Instead we use expert description and classification experiments to verify and investigate their nature.
Age information for individual patients was not available in this study. However, the datasets were composed of patients from multiple clinical studies, for which the average study age was available. The computed mean ages by weighting the mean ages of individual studies can be found in the supplementary material. In principle, a younger age of the healthy group could present a possible confounder in the biomarker identification and evaluation process. However, our data comes from clinical trials with a relatively high mean age (65.9 or higher). Additionally, signs of normal aging in OCT (i.e. mainly retinal pigment epithelium thinning~\cite{ko2017associations}) are less pronounced than AMD related changes. Therefore, we expect that in this study, our method primarily picks up features associated with disease.
A further limitation is that contralateral OCTs of patients with RVO/AMD in the other eye were used as healthy training data. In order to minimize the influence of this potential bias, retina experts conducted careful selection of healthy OCTs within this data.
There is a lack of scientific consensus regarding where normal aging of the retina stops and age-related disease starts. To address this limitation and to account for age-related changes that normally do not result in visual impairment, we have specifically included the mildest category of age-related changes which include small hard drusen ($<63\mu m$)~\cite{ferris2013clinical}.
Another limitation is the restricted informative value of feature importance values in the case of low numbers of examples. A substantially higher number of decision trees is necessary to obtain stable feature scoring results compared to obtaining good classification accuracy.
A further limitation is that the evaluation was conducted with AMD cases only, but, since the applicability of the proposed method is not limited to a specific anomaly, an extension to other diseases should be straightforward.

\section{Conclusion}
We propose a method to segment anomalies in OCT volumes and categorize these findings into disease marker candidates. The detection of new anomalies, rather than the automation of expert annotation of known anomalies, is a critical shift in medical image analysis and particularly relevant in retinal imaging. In this context, we introduced a novel way to identify biomarker candidates, where results on both classification tasks indicate that valuable discriminative information is encoded in the newly identified categories. Future work is needed to transform these categories from candidates to actual markers applicable in clinical practice.

\ifCLASSOPTIONcaptionsoff
  \newpage
\fi



\bibliographystyle{IEEEtran}
\bibliography{references}

%








\end{document}